\documentclass[10pt,twocolumn,letterpaper]{article}

\usepackage[pagenumbers]{cvpr} 

\usepackage[dvipsnames]{xcolor}

\definecolor{cvprblue}{rgb}{0.21,0.49,0.74}
\usepackage[pagebackref,breaklinks,colorlinks,citecolor=cvprblue]{hyperref}

\def\MethodName{StereoNeRF}

\usepackage[accsupp]{axessibility}
\usepackage{kotex}
\usepackage{graphicx}
\usepackage[capitalize]{cleveref}
\Crefname{section}{Sec.}{Secs.}
\Crefname{section}{Section}{Sections}
\Crefname{table}{Table}{Tables}
\crefname{table}{Tab.}{Tabs.}
\usepackage{mathtools}
\usepackage{framed}
\usepackage[table]{colortbl}

\def\paperID{9536} 
\def\confName{CVPR}
\def\confYear{2024}

\title{Generalizable Novel-View Synthesis using a Stereo Camera}

\author{Haechan Lee$^{1,\ast}$ \and Wonjoon Jin$^{2,\ast}$ \and Seung-Hwan Baek$^{1,2}$ \and Sunghyun Cho$^{1,2}$\\
\vspace{-4.8mm}
\and
POSTECH $^1$GSAI \& $^2$CSE\\
{\tt\small \{gocks8, jinwj1996, shwbaek, s.cho\}@postech.ac.kr}\\
\tt\small\href{https://jinwonjoon.github.io/stereonerf/}{jinwonjoon.github.io/stereonerf}
\vspace{-4mm}
}

\begin{document}
\maketitle
\vspace{-4mm}
{\let\thefootnote\relax\footnotetext{\noindent ${}^{\ast}$Equal contribution.}

\begin{abstract}
In this paper, we propose the first generalizable view synthesis approach that specifically targets multi-view stereo-camera images.
Since recent stereo matching has demonstrated accurate geometry prediction, we introduce stereo matching into novel-view synthesis for high-quality geometry reconstruction.
To this end, this paper proposes a novel framework, dubbed \MethodName{}, which integrates stereo matching into a NeRF-based generalizable view synthesis approach.
\MethodName{} is equipped with three key components to effectively exploit stereo matching in novel-view synthesis: a stereo feature extractor, a depth-guided plane-sweeping, and a stereo depth loss.
Moreover, we propose the StereoNVS dataset, the first multi-view dataset of stereo-camera images, encompassing a wide variety of both real and synthetic scenes.
Our experimental results demonstrate that \MethodName{} surpasses previous approaches in generalizable view synthesis.

\end{abstract}    
\vspace{-4mm}

\section{Introduction}
\label{sec:intro}

Novel-view synthesis is a long-standing ill-posed problem in computer vision and graphics, which is inherently challenging due to the necessity of predicting both the geometry and texture from images of a target scene.
Recently, Neural Radiance Fields (NeRF)~\cite{mildenhall2020nerf} have achieved photorealistic results by jointly optimizing geometry and radiance fields with a coordinate-based network. 
However, the need for per-scene optimization, which has to learn representations for each target scene individually, restricts its applicability, as it requires additional training time for such optimization.

\begin{figure}[t!]
\centering
\includegraphics[width=\linewidth]{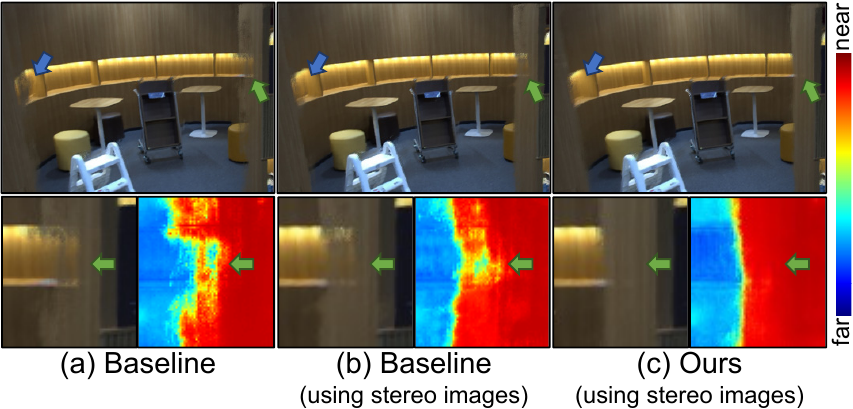}
\vspace{-6mm}
\caption{
Novel-view synthesis results of a baseline method~\cite{johari2022geonerf} and ours.
The baseline shows degraded performances, even trained using stereo-camera images (b).
In contrast, fully exploiting stereo-camera images, our method shows superior results (c).
}
\label{fig:teaser}
\vspace{-6mm}
\end{figure}

Recent approaches~\cite{wang2021ibrnet,yu2021pixelnerf,chibane2021stereo,chen2021mvsnerf,johari2022geonerf,liu2022neural,suhail2022generalizable,varma2022attention} have addressed this issue of synthesizing novel-view images on-the-fly for unseen scenes without per-scene optimization. 
Early studies~\cite{wang2021ibrnet,yu2021pixelnerf,chibane2021stereo} utilize an image encoder to train a generic view interpolation function, enabling the estimation of NeRF parameters from unseen input images in a feed-forward manner. 
However, this single feed-forward manner for estimating geometry and color exacerbates the ill-posedness, resulting in low-quality geometries and rendering results.
For better geometry reasoning, MVSNeRF~\cite{chen2021mvsnerf} and GeoNeRF~\cite{johari2022geonerf} leverage the multi-view stereo (MVS) approach to handle occlusions in a 3D scene.
Nonetheless, they struggle with inaccurate geometry prediction and limited synthesis accuracy, as shown in the synthesis result of GeoNeRF~(\cref{fig:teaser} (a)).

To tackle this challenge, we propose the first generalizable NeRF approach that leverages \textit{stereo-camera} images, which are easily accessible thanks to the ubiquity of stereo cameras in most mobile devices.
Recent advance in learning-based stereo estimation has demonstrated accurate geometry prediction, often even outperforming learning-based MVS methods as shown in~\cref{fig:mvs_stereo}.
The superior performance of stereo estimation can be attributed to several key factors.
First, unlike MVS that assumes arbitrary number of inputs with arbitrary camera positions and orientations, stereo matching assumes two stereo inputs with a fixed baseline.
This constraint allows a more optimal network architecture that can effectively find dense correspondences between input images, and significantly eases the training difficulty. 
Moreover, larger scale of stereo-matching datasets~\cite{mayer2016large,scharstein2002taxonomy,geiger2012we} compared to MVS datasets has facilitated remarkable generalization capabilities in stereo estimation network.
Therefore, we aim to harness this accurate geometric information from stereo images to alleviate the ill-posedness of generalizable view synthesis.
However, since previous methods do not explicitly consider stereoscopic inputs, they cannot leverage the aforementioned benefits, resulting in degenerated performances shown in~\cref{fig:teaser}~(b) despite using stereo-camera images.

\begin{figure}[t!]
\centering
\includegraphics[width=\linewidth]{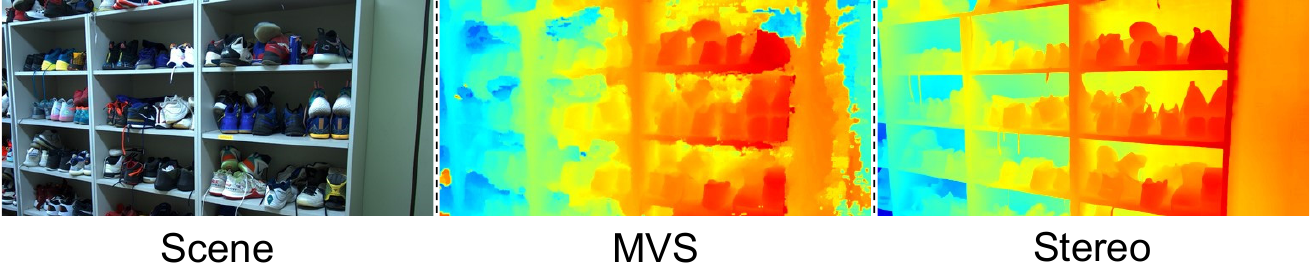}
\vspace{-6mm}
\caption{
Usefulness of exploiting binocular stereo. Comparison on depth estimation between a learning-based MVS method~\cite{peng2022rethinking} and a learning-based binocular stereo method~\cite{xu2023unifying}. 
}
\vspace{-5.3mm}
\label{fig:mvs_stereo}
\end{figure}

This paper proposes \MethodName{}, a novel generalizable view synthesis framework leveraging stereo images.
\MethodName{} integrates stereo matching into NeRF-based generalizable view synthesis approach, where the stereo matching provides vital geometric information.
To this end, we first introduce a stereo feature extractor, which extracts geometry-aware features by correlating horizontal epipolar lines within stereo images.
In addition, the stereo feature extractor takes advantage of stereo-correlated features from an off-the-shelf stereo estimation network, which can transfer rich geometric knowledge to our model.
Furthermore, with the reliable depth estimated from the stereo estimation network, we aggregate multi-view features through a depth-guided plane-sweeping technique.
This technique ensures correspondence matching around the geometry in cost volume construction.
We also present a stereo depth loss utilizing the estimated stereo depth.
These additional geometric cues from the stereo matching effectively mitigate the ill-posedness in generalizable view synthesis.
Notably, our framework leveraging stereo images surpasses the previous approaches~\cite{johari2022geonerf,liu2022neural}, which rely on extra depth information.

Furthermore, we propose the StereoNVS dataset, which is the first dataset for training and evaluation of novel-view synthesis using stereo-camera images.
Our StereoNVS dataset provides real-world and synthetic stereo images.
Our extensive evaluation on the StereoNVS dataset shows that stereo-pair inputs can effectively enhance the quality of novel-view synthesis, and shows that \MethodName{} outperforms previous generalizable novel-view synthesis approaches in terms of image and shape qualities.

Our contributions are as follows:
\begin{itemize}
    \item We propose a generalized NeRF approach that leverages stereo-camera images for the first time to alleviate the ill-posedness in generalizable novel-view synthesis.
    \item We propose a novel framework, \MethodName{}, which exploits the benefits of stereo images by integrating stereo matching into generalizable view synthesis. To this end, we present a stereo feature extractor, a depth-guided plane-sweeping, and a stereo depth loss in our framework.
    \item We also present the StereoNVS dataset, the first dataset for training and evaluation of novel-view synthesis obtained by stereo cameras.
\end{itemize}

\section{Related Work}
\label{sec:related_work}

\subsection{Novel-View Synthesis}
Novel-view synthesis aims to synthesize target-view images from reference-view images.
Early approaches synthesize novel-view images by blending pixels from multiple input images~\cite{debevec1996modeling,levoy1996light,gortler1996lumigraph}.
Recent work adopting neural volume representations has shown remarkable novel-view synthesis results.
Zhou et al.~\cite{zhou2018stereo} propose multi-plane images (MPI) representation estimated from input images, but their methods produce valid novel-view images only for narrow ranges of camera poses.
Mildenhall et al.~\cite{mildenhall2020nerf} propose Neural Radiance Fields (NeRF) that can synthesize photo-realistic target-view images via neural implicit representations and volume rendering.
Albeit its photo-realism, computation-heavy per-scene optimization is needed.
Recent variants such as~\cite{fridovich2022plenoxels,yu2021plenoctrees,muller2022instant,kerbl20233d} have remarkably reduced the optimization time, but they still require large memory and several minutes for training.

\vspace{-3mm}
\paragraph{Generalizable View Synthesis.}
Recently, many novel-view synthesis approaches without per-scene optimization have been proposed. 
Several studies directly predict pixel colors by aggregating image features from aligned pixels, without 3D representations.
Suhail et al.~\cite{suhail2022generalizable} and Varma et al.~\cite{varma2022attention} adopt a transformer-based network~\cite{dosovitskiy2020image} to compute features along the epipolar lines, which needs a large number of training images for high-quality view synthesis.
Du et al.~\cite{du2023learning} propose a framework to synthesize target views from two images with small overlapped regions.

Another research direction predicts volumetric representation~\cite{mildenhall2020nerf} from aggregated features from reference views and synthesizes images via volumetric rendering.
PixelNeRF, SRF, and GRF~\cite{yu2021pixelnerf,chibane2021stereo,trevithick2021grf} predict radiance fields from pixel-aligned features using an MLP.
Among these methods, SRF~\cite{chibane2021stereo} uses two sampled images from an image collection captured by a monocular camera as a stereo pair, but shows limitied synthesis quality.
IBRNet~\cite{wang2021ibrnet} proposes to learn generic view interpolation functions, but suffers from artifacts for challenging scenes with complex geometries.
MVSNeRF, GeoNeRF and NeuRay~\cite{chen2021mvsnerf,johari2022geonerf,liu2022neural} utilize the MVS approach using cost volume for better occlusion handling in generalizable view synthesis.

However, all the aforementioned methods often fail to capture accurate geometry, particularly in textureless regions, leading to limited synthesis results.
Unlike these methods, our framework exploits stereo-camera images with a fixed baseline to effectively capture the geometry of complex scenes as well as textureless regions for high-quality novel-view synthesis.

\subsection{Stereo Matching and Multi-View Stereo}
Geometry estimation is a long-standing problem in computer vision that has a variety of applications. 
Among them, stereo matching is a task that takes rectified stereo images and computes stereo correspondence to estimate disparities~\cite{scharstein2002taxonomy,hartley2003multiple}.
Recently, a huge number of data-driven approaches have been introduced with the emergence of a vast amount of stereo-matching datasets~\cite{mayer2016large,scharstein2002taxonomy,geiger2012we} and made significant progress~\cite{xu2023unifying,chang2018pyramid,zhang2019ga,mayer2016large}.
MVS approaches that use more than two views have been extensively studied as well~\cite{furukawa2015multi,seitz1999photorealistic,kutulakos2000theory}.
Recently, learning-based MVS approaches have been proposed, e.g., Cheng et al.~\cite{cheng2020deep}, Gu et al.~\cite{gu2020cascade}, and Yang et al.~\cite{yang2020cost} present efficient frameworks that cascade cost volumes in a coarse-to-fine manner to enable high-resolution depth estimation.

In our work, we have the best of both worlds in our generalizable view synthesis framework.
We adopt the MVS approach to aggregate multi-view information for occlusion-aware geometry estimation, as done in GeoNeRF~\cite{johari2022geonerf}.
Furthermore, our framework also assumes structured stereo images as inputs, and integrates the two-view stereo matching into our framework for robust and accurate geometry estimation in generalizable novel-view synthesis.

\section{\MethodName{}}
\label{sec:method}

For novel-view synthesis, our framework takes a set of rectified stereo-camera images and estimates neural radiance fields~\cite{mildenhall2020nerf} from which novel-view images are rendered.
\MethodName{} integrates a well-designed stereo-matching algorithm into the existing generalizable view synthesis approach~\cite{johari2022geonerf}.
In this section, we first provide an overall pipeline that utilizes an pre-trained stereo estimation network~\cite{xu2023unifying}.
Then, we explain each step of our framework and the training process in detail, highlighting how to integrate stereo matching into our framework.

\vspace{-2mm}
\paragraph{Overall pipeline.}
\cref{fig:overview} shows an overview of our pipeline.
Our framework utilizes $N$ pairs of stereo images of a target scene $\{(I_L^n, I_R^n)\}_{n=1}^{N}$ for novel-view synthesis.
In the first step, the pre-trained stereo estimation network takes the $n$-th image pair $(I_L^n, I_R^n)$, and outputs stereo depths $(d_{s,L}^n,d_{s,R}^n)$ and stereo-correlated features $(t_L^n,t_R^n)$.
In the second step, a stereo feature extractor takes the stereo image pair and the stereo-correlated features, and outputs stereo image features $(f_L^n,f_R^n)$, explained in~\cref{sec:feat_ext_img}.
The third step aggregates stereo image features from all viewpoints $\{(f_L^n,f_R^n)\}_{n=1}^{N}$ to build 3D feature volumes $\{(\phi_L^n,\phi_R^n)\}_{n=1}^{N}$ using an MVS network~(\cref{sec:feat_ext_vol}).
For constructing feature volumes that faithfully reflect the 3D geometry of a target scene, the third step adopts a depth-guided plane-sweeping which utilizes the estimated stereo depths $(d_{s,L}^n,d_{s,R}^n)$.
Finally, a neural renderer predicts radiance fields from the stereo image features $\{(f_L^n,f_R^n)\}_{n=1}^{N}$ and the feature volumes $\{(\phi_L^n,\phi_R^n)\}_{n=1}^{N}$ from all viewpoints, and novel-view images are synthesized from the radiance fields~(\cref{sec:rendering}).

\begin{figure}[t!]
\centering
\includegraphics[width=\linewidth]{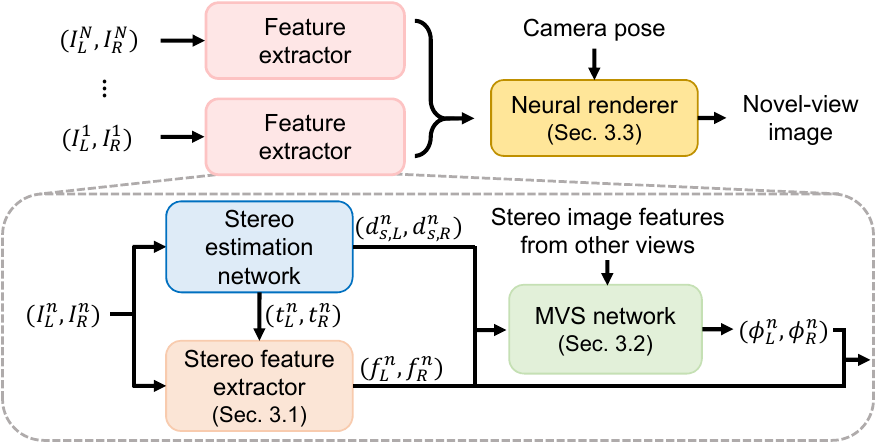}
\caption{
Overview of \MethodName{}. \MethodName{} consists of a shared feature extractor and a neural renderer. We design the feature extractor with a stereo estimation network, a stereo feature extractor, and a MVS network. \MethodName{} takes $N$ pairs of stereo images and a camera pose as inputs, and synthesizes a novel-view image of the camera pose.
}
\label{fig:overview}
\vspace{-4mm}
\end{figure}

\subsection{Feature Extraction from Stereo Image Pairs}
\label{sec:feat_ext_img}
Unlike previous approaches, which extract features of input images respectively, our stereo feature extractor takes rectified stereo-camera images and computes image feature maps by exploiting the epipolar geometry between them.
To this end, our stereo feature extractor consists of three parts: CNN encoders, stereo attention modules (SAM)~\cite{chu2022nafssr}, and CNN decoders (\cref{fig:feat_ext}).
Moreover, we propose integrating the stereo-correlated features $(t_L^n,t_R^n)$ from the pre-trained stereo estimation network into the SAM (green line in~\cref{fig:feat_ext}).
This integration inherits geometric cues from the stereo estimation network.
In the following, we provide the detailed description of the stereo feature extractor.

\begin{figure}[t!]
\centering
\includegraphics[width=\linewidth]{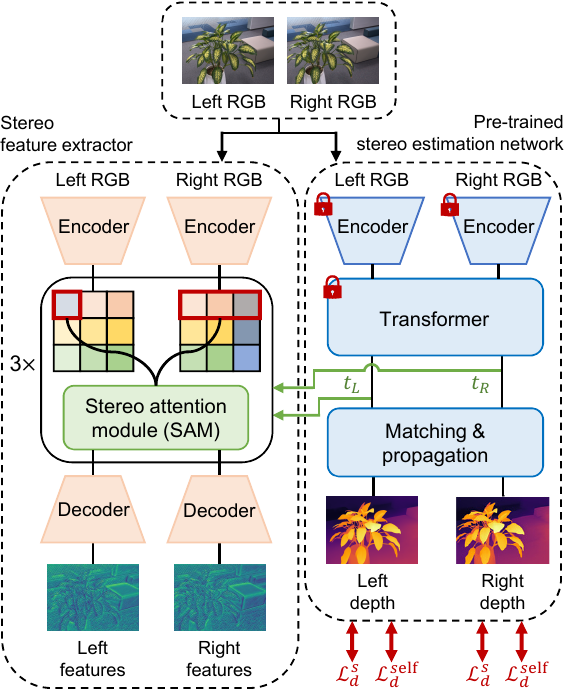}
\caption{
Stereo feature extractor of \MethodName{}, where an pre-trained stereo estimation network is incorporated.
}
\label{fig:feat_ext}
\vspace{-4mm}
\end{figure}

First, the weight-shared CNN encoders project each of $I_L^n$ and $I_R^n$ into the feature space.
Then, we adopt the stereo attention module (SAM)~\cite{chu2022nafssr} to fuse two feature maps on the horizontal epipolar lines.
To this end, SAM estimates stereo correspondences between two feature maps and explicitly adds stereo-correspondent features, as shown in~\cref{fig:sam}.
In addition, we extend the SAM by aggregating the stereo-correlated features $(t_L^n, t_R^n)$ in this feature fusion process.
Note that the stereo-correlated features are computed from the pre-trained stereo estimation network, which provides vital geometric cues.
These stereo-correlated features enhance stereo correspondences between two feature maps, leading to higher-quality fused feature maps.

Specific explanation of the aforementioned feature fusion of the SAM is as follows~(\cref{fig:sam}). For the left feature, a warping matrix $\mathcal{W}_{R \rightarrow L}$ is built by multiplying a query matrix and a key matrix followed by a softmax layer.
The query matrix is computed from the left feature and the stereo-correlated features, and the key matrix is computed from the right feature and the stereo-correlated features.
Then, a value matrix, computed from the right feature and the stereo-correlated feature, is warped along the epipolar lines by multiplying the warping matrix $\mathcal{W}_{R \rightarrow L}$, then added to the left feature.
The same feature fusion process is also performed for the right feature.

For effective fusion of information from both stereo images and stereo-correlated features, the stereo feature extractor has three sequentially stacked SAM.
Finally, the weight-shared CNN decoders take each of the fused features from the SAM, and generate stereo image features $(f_L^n,f_R^n)$, which will be used for building feature volumes.

\begin{figure}[t!]
\centering
\begin{tabular}{@{}c}
\includegraphics[width=\linewidth]{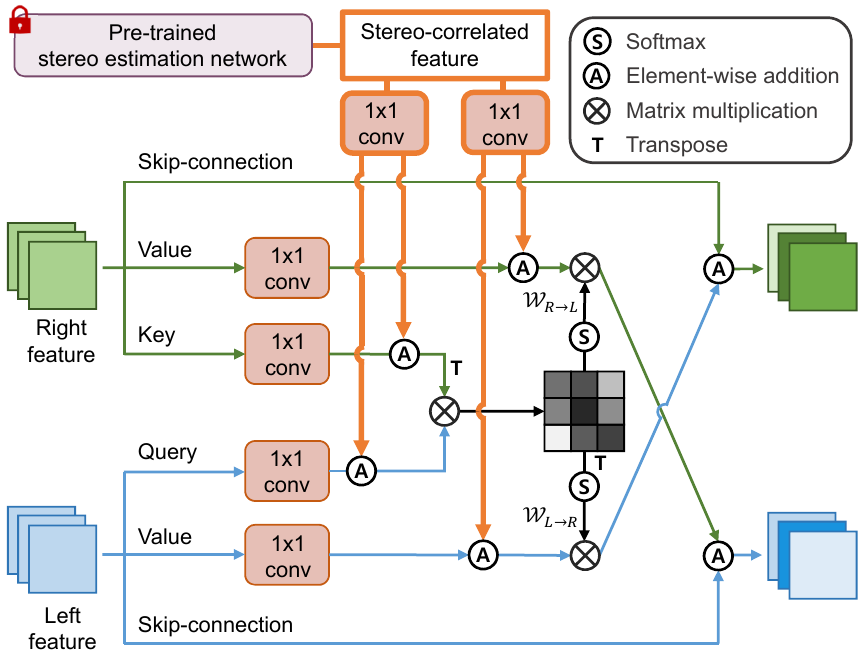} \\
\end{tabular}
\vspace{-2mm}
\caption{
Stereo attention module used in the stereo feature extractor. We exploit the rich features from the pre-trained stereo estimation module.
}
\label{fig:sam}
\vspace{-4mm}
\end{figure}

\subsection{Depth-guided Feature Volume Construction}
\label{sec:feat_ext_vol}
Once stereo image features are obtained, we aggregate these features from all viewpoints to create cost volumes.
To this end, we adopt a plane-sweeping-based approach, which computes multi-view correspondences among these features.
The plane-sweeping-based approach first defines a depth range within pre-computed near and far depths obtained from COLMAP~\cite{schonberger2016structure}.
Then, it hypothesizes depth planes in the entire depth range.
On these depth planes, multi-view stereo image features are aggregated via plane-sweeping to build cost volumes.
Then, these cost volumes are processed to yield feature volumes, which will be used later in predicting neural radiance fields.

However, this plane-sweeping-based approach often struggles with accurate geometry estimation.
Since this approach searches the entire depth range of a scene, correspondence matching is inaccurate and prone to error, especially in textureless regions.
To tackle this issue, we introduce a depth-guided plane-sweeping (DGPS), which constructs cost volumes around the stereo depth predicted from the stereo estimation network.
Unlike the previous approach using the entire depth range, DGPS introduces a dynamic search range that varies based on the estimated stereo depth.
This technique significantly constrains the search space and reduces outliers in correspondence matching.

\cref{fig:feature_volume} depicts the feature volume construction using DGPS. Instead of searching the entire depth range, we hypothesize depth planes around the stereo depths $(d_{s,L}^n,d_{s,R}^n)$.
Then, through DGPS, we aggregate the stereo image features from all viewpoints $\{(f_L^{n},f_R^{n})\}_{n=1}^{N}$ on these depth planes to build cost volumes.
These cost volumes are further processed via the MVS network to obtain feature volumes $(\phi_L^{n},\phi_R^{n})$ and depth maps $(d_{s,L}^n,d_{s,R}^n)$.
We repeat this process to obtain feature volumes and depth maps for every viewpoint.
The resulting multi-view feature volumes and depth maps are denoted as $\{(\phi_L^{L},\phi_R^{L})\}_{n=1}^{N}$ and $\{(d_{m,L}^{n},d_{m,R}^{n})\}_{n=1}^{N}$, respectively.
Refer to Sec.~B.3 in the supplementary document for more details about our feature volume construction.

\begin{figure}[t!]
\centering
\includegraphics[width=0.95\linewidth]{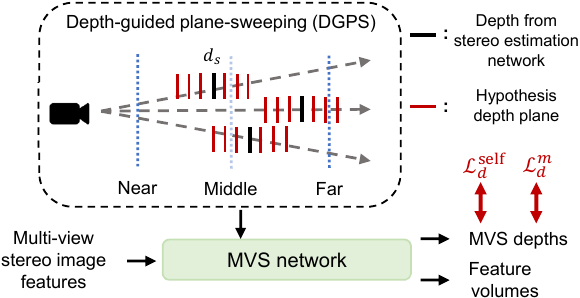}
\vspace{-2mm}
\caption{Feature volume construction using DGPS. 
}
\label{fig:feature_volume}
\vspace{-3mm}
\end{figure}

\subsection{Rendering Novel Views}
\label{sec:rendering}
Once feature volumes are constructed, novel-view images can be rendered via neural rendering.
For rendering novel-view images, our framework adopts the rendering procedure of GeoNeRF~\cite{johari2022geonerf}.
In the following, we briefly describe the rendering procedure.
Given a target viewpoint, we cast a ray for each pixel and sample points along the ray in the 3D space.
Then, for each sampled point, we sample features from all the feature volumes $\{(\phi_L^{n},\phi_R^{n})\}_{n=1}^{N}$.
We also sample features from all the image features $\{(f_L^{n},f_R^{n})\}_{n=1}^{N}$ by projecting the sampled point onto input images.
The sampled volume and image features are aggregated by a neural renderer network, and the color and density are estimated.
Finally, by integrating the estimated color values and depth values of sampled points with their densities along each ray, we obtain the color $\hat{c}$ and the depth $d_{r}$ at each pixel of a novel view, respectively.

\subsection{Training \MethodName{}}
\label{sec:loss}
We train \MethodName{} with ground-truth images and pseudo-ground-truth depths for high-quality novel-view synthesis.
The pseudo-ground-truth depths $d_{gt}$ are obtained from the pre-trained stereo estimation network~\cite{xu2023unifying}. 

Our training loss $\mathcal{L}$ is defined as follows:
\begin{equation}
\begin{aligned}
\mathcal{L} = \mathcal{L}_{c} + \mathcal{L}_{d},
\end{aligned}
\end{equation}
where $\mathcal{L}_{c}$ is a color loss and $\mathcal{L}_{d}$ is a depth loss.
The color loss is defined as a mean-squared-error (MSE) between the rendered colors $\hat{c}$ and ground-truth colors~\cite{mildenhall2020nerf}.

To alleviate the shape-radiance ambiguity~\cite{zhang2020nerf++} in scene reconstruction using neural radiance fields, we introduce the depth loss $\mathcal{L}_{d}$, which is defined as:
\begin{equation}
\begin{aligned}
\mathcal{L}_{d} = \lambda_{d}^{self}\mathcal{L}_{d}^{self} + \lambda_{d}^{stereo}\mathcal{L}_{d}^{stereo},
\end{aligned}
\end{equation}
where $\mathcal{L}_{d}^{self}$ is a self-supervised depth loss from GeoNeRF~\cite{johari2022geonerf} and $\mathcal{L}_{d}^{stereo}$ is our proposed stereo depth loss.
$\mathcal{L}_{d}^{self}$ penalizes the depths estimated from both the stereo estimation network ($d_{s}$) and the MVS network ($d_{m}$) by comparing them to the depth rendered from NeRF ($d_{r}$).

Our framework estimates depths from the stereo estimation network ($d_{s}$), the MVS network ($d_{m}$), and the neural renderer network ($d_{r}$).
The stereo depth loss $\mathcal{L}_{d}^{stereo}$ guides these networks to predict more accurate depths using the pseudo-ground-truth depths $d_{gt}$.
This loss introduces a further quality improvement, especially in geometric details as will be shown in \cref{sec:analysis}.
$\mathcal{L}_{d}^{stereo}$ is defined as:
\begin{equation}
\begin{aligned}
\label{equ:stereo_depth}
\mathcal{L}_{d}^{stereo} = \lambda_{1}\mathcal{L}_{d}^{s} + \lambda_{2} \mathcal{L}_{d}^{m} + \lambda_{3} \mathcal{L}_{d}^{r},
\end{aligned}
\end{equation}
where $\mathcal{L}_{d}^{s}$, $\mathcal{L}_{d}^{m}$, and $\mathcal{L}_{d}^{r}$ penalizes depths $d_{s}$, $d_{m}$, and $d_{r}$, respectively, by comparing them with $d_{gt}$.
Note that $d_{gt}$ is different from $d_{s}$. We obtain $d_{gt}$ from the pre-trained stereo estimation network with frozen parameters. 
On the other hand, we obtain $d_{s}$ from the stereo estimation network in our framework, which has trainable parameters.
Refer to Sec.~B.4 in the supplementary document for more details about our stereo depth loss.

Due to the different characteristics of datasets such as baseline length, $d_s$ may have estimation error, which leads to an error in the depth-guided plane-sweeping.
To tackle this, we partially train the matching and propagation network in the stereo estimation network~(\cref{fig:feat_ext}) using $\mathcal{L}_d^{self}$.
$\mathcal{L}_d^{self}$ provides multi-view supervision to the stereo estimation network by leveraging $d_r$ estimated from multi-view images.
However, since $d_r$ may also have estimation error, we further regularize the stereo estimation network using $\mathcal{L}_d^{s}$.
This training scheme for the stereo estimation network ensures consistent depth estimation, while preserving the stereo matching capability.
Sec.~D.1 in the supplementary document further discusses the training scheme of the stereo estimation network.

\section{StereoNVS Dataset}
\label{sec:data}
We propose the StereoNVS dataset, the first stereo-camera image dataset for training and evaluating novel-view synthesis using stereo-camera images.
The StereoNVS dataset provides both real and synthetic datasets, each of which is dubbed StereoNVS-Real and StereoNVS-Synthetic.
In the following, we present the details of each dataset.

\begin{figure*}[!t]
\centering
\begin{tabular}{@{}c}
\includegraphics[width=\linewidth]{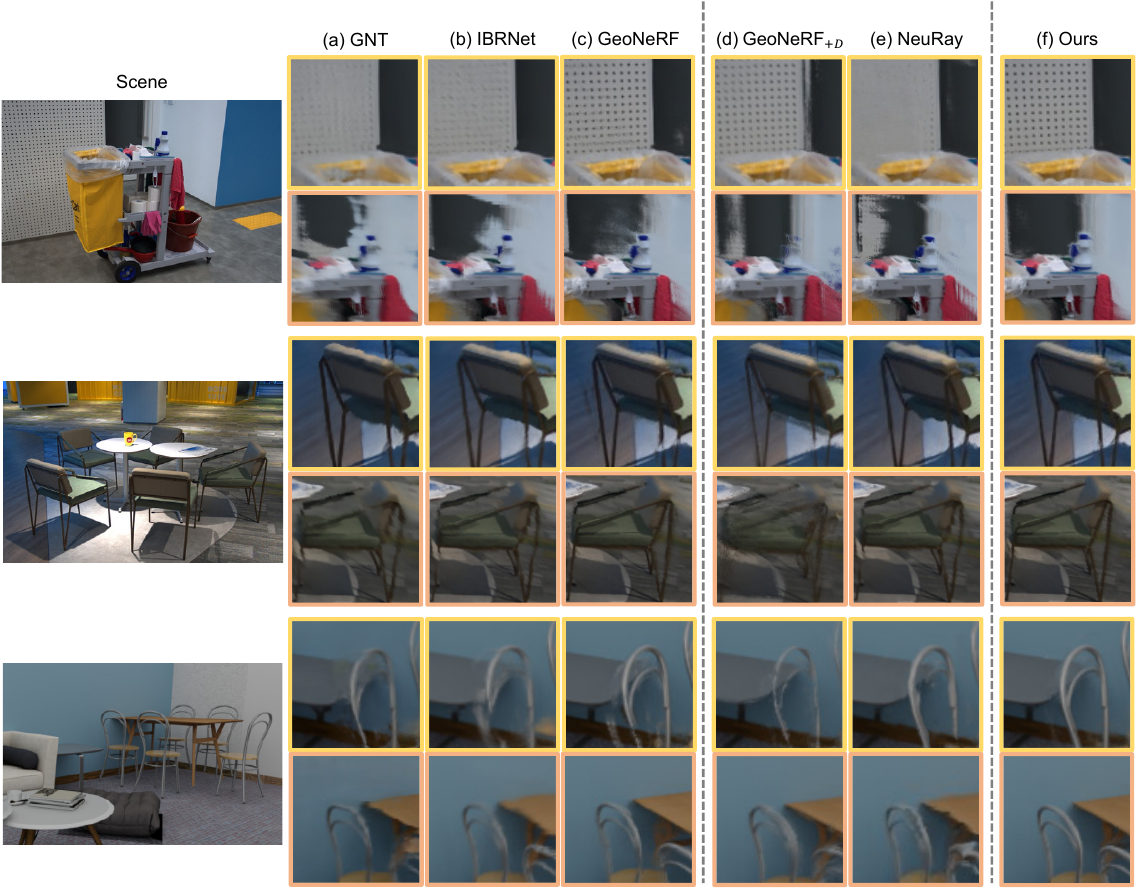} \\
\end{tabular}
\vspace{-4mm}
\caption{
Qualitative comparison of novel-view synthesis on the StereoNVS dataset, showing rendering results for two real-world scenes (above) and one synthetic scene (below).
All models are trained using stereo images.
Our method outperforms the baseline methods~\cite{wang2021ibrnet,varma2022attention,johari2022geonerf,liu2022neural} on both real-world and synthetic scenes, especially in thin structures and textureless regions. 
}
\label{fig:qual_main}
\vspace{-4mm}
\end{figure*}

\subsection{StereoNVS-Real}
StereoNVS-Real provides real-world stereo-camera images for the training and evaluation of novel-view synthesis.
The dataset provides stereo-camera images of 53 static scenes,
and around 25 stereo-image pairs per scene.
The images are undistorted and stereo-rectified, and have a resolution of $1792\times 896$.
The camera parameters such as the camera poses are provided as well.
In our experiments, we divide the dataset into 45 and 8 scenes as training and test sets.

To capture stereo images, we built a camera rig with two Basler machine vision cameras.
We measured the camera parameters including the intrinsic and distortion parameters by camera calibration using multi-view images of a checkerboard~\cite{zhang2000flexible}.
Then, we collected stereo images from multiple viewpoints for various indoor and bounded scenes using our camera system.
The captured images were then undistorted and stereo-rectified.
Finally, we obtained the camera poses of the captured images using COLMAP~\cite{schonberger2016structure}.

\subsection{StereoNVS-Synthetic}
For quantitative evaluation of synthesized geometries, we also present the StereoNVS-Synthetic dataset, constructed by rendering synthetic 3D models using the 3D-Front dataset~\cite{fu20213d}.
The images have a resolution of $864\times 448$, which is half of the real dataset.
StereoNVS-Synthetic provides stereo-camera images of 50 scenes, and around 150 stereo image pairs per scene
as well as the ground-truth camera parameters, camera poses, and depth maps.
For more details about the StereoNVS dataset, refer to Sec.~E in the supplementary document.

\section{Experiments}
\label{sec:exp}

We conduct extensive validation of our method on the StereoNVS dataset.
In the following, we will provide implementation details of our method (\cref{sec:implementation_detail}), compare our method with other baselines (\cref{sec:comparison}), and conduct a comprehensive analysis of our proposed components, which proves the effectiveness of our framework including the stereo feature extractor, the depth-guided plane-sweeping, and the stereo depth loss~(\cref{sec:analysis}).

\begin{table}[!t]
\centering
\resizebox{\columnwidth}{!}{
    \setlength\tabcolsep{1.5pt}
    \begin{tabular}{c|cccc|cccc}
    \toprule[1pt]
        ~  & \multicolumn{4}{c|}{StereoNVS-Real}  & \multicolumn{4}{c}{StereoNVS-Synthetic}  \\ 
        Method & PSNR($\uparrow$) & SSIM($\uparrow$) & LPIPS($\downarrow$) & ABS($\downarrow$) & PSNR($\uparrow$) & SSIM($\uparrow$) & LPIPS($\downarrow$) & ABS($\downarrow$)  \\ \hline
        SRF            & 21.12 & 0.6933 & 0.4164    & 2.9176 & 22.36 & 0.7162 & 0.4245 & 0.8125 \\ 
        IBRNet         & 26.12 & 0.8421 & 0.2095    & 0.7566 & 30.83 & 0.8889 & 0.1823 & 0.2628 \\ 
        GeoNeRF        & \cellcolor{yellow!30}28.01 & \cellcolor{yellow!30}0.8929 & \cellcolor{yellow!30}0.1460 & \cellcolor{yellow!30}0.5064    & 32.13 & 0.9179 & 0.1438 & 0.1577  \\ 
        GNT            & 26.08 & 0.8434 & 0.2285    & 1.0959 & 26.17 & 0.8406 & 0.2650 & 0.4512 \\ \hline 
        GeoNeRF$_{+D}$ & 25.65 & 0.8172 & 0.2082    & 0.8693 & \cellcolor{yellow!30}32.85 & \cellcolor{yellow!30}0.9321 & \cellcolor{yellow!100}0.1171 & \cellcolor{yellow!100}0.0782  \\ 
        NeuRay         & 26.51 & 0.8538 & 0.1887    & 0.6147 & 32.26 & 0.9104 & 0.1478 & 0.1571 \\ \hline 
        Ours           & \cellcolor{yellow!100}28.44 & \cellcolor{yellow!100}0.9000 & \cellcolor{yellow!100}0.1396 & \cellcolor{yellow!100}0.4868   & \cellcolor{yellow!100}33.45 & \cellcolor{yellow!100}0.9336 & \cellcolor{yellow!30}0.1203 & \cellcolor{yellow!30}0.1056 \\ 
    \bottomrule[1pt]
    \end{tabular}
}
\vspace{-3mm}
\caption{Quantitative comparison between the baseline methods~\cite{chibane2021stereo,wang2021ibrnet,varma2022attention,johari2022geonerf,liu2022neural} and ours.
All models are trained using stereo images. 
Our method shows superior performance on both StereoNVS-Real and StereoNVS-Synthetic datasets. 
}
\label{table:comp}
\vspace{-4mm}
\end{table}

\subsection{Implementation Details}
\label{sec:implementation_detail}
We train our model on the training set of StereoNVS-Real, which has multi-view stereo image pairs of real scenes.
Our model is trained for 250K iterations.
For each iteration, we randomly select one scene and one target viewpoint of the scene.
For both training and evaluation, image features and feature volumes are extracted from three stereo image pairs (i.e., total six images) at three viewpoints nearest to the target viewpoint.
During training, 512 rays are randomly selected for the training batch.
We use the Adam optimizer~\cite{kingma2014adam} with learning rates of 0.0005 and the cosine annealing scheduling~\cite{loshchilov2016sgdr}.
We employ UniMatch~\cite{xu2023unifying} for both the pre-trained stereo estimation network within the stereo feature extractor and the pseudo-ground-truth depth of the stereo depth loss.
Refer to Sec.~B in the supplementary document for additional implementation details.

\subsection{Comparison}
\label{sec:comparison}
We compare our method with recent generalizable novel view synthesis methods: SRF~\cite{chibane2021stereo}, IBRNet~\cite{wang2021ibrnet}, GeoNeRF~\cite{johari2022geonerf}, GNT~\cite{varma2022attention} and NeuRay~\cite{liu2022neural}.
Like our method, we use three stereo-camera image pairs to synthesize each target-view image for all the baseline methods.
While the previous methods do not explicitly assume stereo-camera images as their inputs, we also train them using stereo-camera images as they can handle stereo-camera images as independent inputs.
All the baseline models are trained on the training set of StereoNVS-Real, as done for our method.

We evaluate both image and depth qualities on the StereoNVS-Real and the StereoNVS-Synthetic datasets.
For StereoNVS-Real, we utilize pseudo-ground-truth depths obtained from COLMAP~\cite{schonberger2016structure} to assess depth quality. On the other hand, for StereoNVS-Synthetic, we use rendered depth maps as ground truths for depth quality assessment.
We employ PSNR, SSIM~\cite{wang2004image}, and LPIPS~\cite{zhang2018unreasonable} as metrics for image quality and absolute error (ABS) as a metric for depth quality.

Among the compared methods, for training and inference, NeuRay~\cite{liu2022neural} requires depth maps and GeoNeRF~\cite{johari2022geonerf} can use depth maps as additional inputs.
We denote such variation of GeoNeRF as GeoNeRF$_{+D}$.
For their training and inference, we used the pseudo-ground-truth depth maps estimated by UniMatch~\cite{xu2023unifying} as done for our method.

\vspace{-2.3mm}
\paragraph{Qualitative comparison.}
\cref{fig:qual_main} presents a qualitative comparison on novel view synthesis between our method and previous methods~\cite{wang2021ibrnet,varma2022attention,johari2022geonerf,liu2022neural} using the StereoNVS dataset.
Previous methods estimate inaccurate geometry for scenes with thin structures or textureless regions, resulting in severe artifacts in the synthesized novel view images.
In contrast, our method clearly outperforms the baseline methods in view synthesis results with significantly fewer artifacts even in textureless regions, thanks to more accurately estimated geometry.
Moreover, our method shows better synthesis results compared to GeoNeRF$_{+D}$~\cite{johari2022geonerf} and NeuRay~\cite{liu2022neural}, even though they explicitly use depth maps in the inference time, demonstrating the robust utilization of depth maps in our framework.

\vspace{-2mm}
\paragraph{Quantitative comparison.}
\cref{table:comp} shows that our method generally surpasses other baseline methods~\cite{chibane2021stereo,wang2021ibrnet,varma2022attention,johari2022geonerf,liu2022neural} in terms of image and depth qualities.
GNT~\cite{varma2022attention} exhibits degraded performances, likely due to its data-hungry transformer backbone.
While performance differences with GeoNeRF are not substantial on the StereoNVS-Real dataset, our method shows the superior perceptual quality~(\cref{fig:qual_main}) and significantly better performance on the StereoNVS-Synthetic dataset.
While GeoNeRF$_{+D}$~\cite{johari2022geonerf} and NeuRay~\cite{liu2022neural} achieve comparable performances on the StereoNVS-Synthetic dataset, they show considerably degenerated results on the StereoNVS-Real dataset.
In contrast, our method demonstrates superior performance thanks to the robustness of our framework.
Sec.~D.1 in the supplementary document further discusses the sensitivity of GeoNeRF$_{+D}$, NeuRay, and our method to depth errors in real-world images.

\subsection{Analysis and Discussion}
\label{sec:analysis}

\begin{table}[!t]
\centering
\resizebox{\linewidth}{!}{
    \setlength\tabcolsep{1.0pt}
    \begin{tabular}{l|cccc|cccc}
    \toprule[1pt]
        ~  & \multicolumn{4}{c|}{StereoNVS-Real}  & \multicolumn{4}{c}{StereoNVS-Synthetic}  \\
        ~ & PSNR($\uparrow$) & SSIM($\uparrow$) & LPIPS($\downarrow$) & ABS ($\downarrow$) & PSNR($\uparrow$) & SSIM($\uparrow$) & LPIPS($\downarrow$) & ABS ($\downarrow$) \\ \hline
         Baseline (a)             & 28.05 & 0.8953 & 0.1372    & 0.5598 & 32.22 & 0.9172 & 0.1386 & 0.1679 \\ 
         + Stereo setting (b)     & 28.01 & 0.8929 & 0.1460    & 0.5064 & 32.13 & 0.9179 & 0.1438 & 0.1577 \\ 
         + SAM (c)                & 28.21 & 0.8967 & 0.1398    & 0.4935 & 31.90 & 0.9167 & 0.1386 & 0.1531 \\ 
         + Correlated feature (d) & 28.31 & 0.8988 & \cellcolor{yellow!100}0.1370 & 0.5057    & 32.35 & 0.9245 & 0.1282 & 0.1416 \\ 
         + DGPS (e)               & 28.42 & 0.8997 & 0.1403    & 0.5098 & 33.09 & 0.9304 & 0.1230 & 0.1246 \\ 
         + Stereo depth loss (f)  & \cellcolor{yellow!100}28.44 & \cellcolor{yellow!100}0.9000 & 0.1396  & \cellcolor{yellow!100}0.4868   & \cellcolor{yellow!100}33.45 & \cellcolor{yellow!100}0.9336 & \cellcolor{yellow!100}0.1203 & \cellcolor{yellow!100}0.1056 \\  
    \bottomrule[1pt]
    \end{tabular}
}
\vspace{-3mm}
\caption{Quantitative ablation study.
}
\label{table:abl_key}
\vspace{-3mm}
\end{table}

\begin{figure}[!t]
\centering
\begin{tabular}{@{}c}
\includegraphics[width=\linewidth]{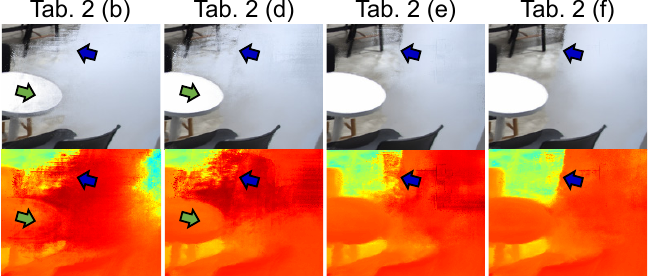} \\
\end{tabular}
\vspace{-3mm}
\caption{
Qualitative ablation study. Our full method enables accurate synthesis of a novel-view image and a depth map. 
}
\label{fig:abl_qual}
\vspace{-4mm}
\end{figure}

\subsubsection{Ablation Study}
To assess the impact of our proposed methods, we conduct an ablation study starting with our baseline model~\cite{johari2022geonerf}, which is trained on three views in a monocular setting (i.e., three images).
Although equipped with a stereo camera setting, the baseline model trained on three stereo views (i.e., six images) shows similar image qualities as shown in~\cref{table:abl_key} (b). This result indicates that sophisticated methods are needed to exploit the invaluable information from stereo-camera images.

\cref{table:abl_key} shows that our model consistently demonstrates improvements in view synthesis results, as we introduce our proposed components.
Our stereo feature extractor enables us to estimate better geometry, leading to fewer artifacts in novel view synthesis, as shown in~\cref{fig:abl_qual} (d).
The stereo attention module (c) and the stereo-correlated features (d) help extract robust stereo image features, which are particularly effective for real scenes and synthetic scenes, respectively.
Our DGPS is essential for better geometry estimation, leading to significantly improved results as shown in~\cref{fig:abl_qual} (e), especially in textureless regions.
This is further evident in~\cref{table:abl_key} (e), with considerable performance gain on the StereoNVS-Synthetic dataset.
Our final model shows the best performances (\cref{table:abl_key} (f)), with high-quality depth and view synthesis results (\cref{fig:abl_qual} (f)).

\vspace{-2.5mm}
\subsubsection{Effectiveness of Depth-Guided Plane-Sweeping}
We conduct two experiments to demonstrate the effectiveness of our DGPS.
In the first experiment, we compare our final model against a model without using DGPS. The ``Final model'' in~\cref{table:abl_sweeping_remove} denotes our final model with DGPS. As shown in~\cref{table:abl_sweeping_remove}, the model without using DGPS shows degenerated results compared to our final model, highlighting the effectiveness of DGPS.

In the second experiment, we compare our model using DGPS against a model using more depth planes for cost volume construction. The term "Base model" in \cref{table:abl_sweeping_add} refers to our model that is solely equipped with the stereo feature extractor, excluding DGPS and the stereo depth loss. We train an additional model that uses approximately 1.5 times the number of depth planes, compared to the base model.
Although this additional model utilizes more depth planes, it shows similar results to the base model on StereoNeRF-Synthetic, as reported in~\cref{table:abl_sweeping_add}.
This is because increasing the number of planes does not guarantee accurate correspondence matching, especially in textureless regions.
In contrast, our DGPS guarantees correspondence matching across multi-view image features near the geometry, resulting in better image and depth qualities.
Refer to Sec.~C.2.1 in the supplementary document for more details. 

\vspace{-2.5mm}
\subsubsection{Benefit of Stereo Estimation in Depth Loss}

To show the effectiveness of using stereo estimation networks for depth supervision, we conduct an additional experiment as follows.
First, we obtain two pseudo-GT depths: one from the pre-trained stereo network~\cite{xu2023unifying} ($d_{gt}$) as stated in~\cref{sec:loss} and the other from the state-of-the-art learning-based MVS network ($d_{gt}^{mvs}$)~\cite{peng2022rethinking}.
Then, we train two models with our methods using $d_{gt}$ and $d_{gt}^{mvs}$, respectively.
Then, we compare their synthesis results based on image and shape qualities.

As shown in~\cref{table:abl_depth}, our model trained with $d_{gt}$ surpasses the other model using $d_{gt}^{mvs}$ on StereoNVS-Synthetic.
Note that the MVS network takes more images than the stereo network, seven images and two images, respectively.
These results show that the stereo network provides a reliable depth signal for high-quality view synthesis thanks to its generalization ability, which came from standardized inputs of stereo-camera images and large-scale stereo datasets.

\begin{table}[!t]
\centering
\resizebox{\columnwidth}{!}{
    \setlength\tabcolsep{1.5pt}
    \begin{tabular}{l|cccc}
    \toprule[1pt]
         & PSNR($\uparrow$) & SSIM($\uparrow$) & LPIPS($\downarrow$) & ABS($\downarrow$) \\ \hline
         Final model (\cref{table:abl_key} (f))           & \textbf{33.45} & \textbf{0.9336} & \textbf{0.1203} & \textbf{0.1056} \\
         \hspace{3mm} - DGPS                              & 32.30 & 0.9240 & 0.1290 & 0.1300 \\
         
    \bottomrule[1pt]
    \end{tabular}
}
\vspace{-2mm}
\caption{
Effectiveness of our depth-guided plane-sweeping (DGPS) compared to the baseline models without using DGPS.
}
\vspace{-2mm}
\label{table:abl_sweeping_remove}
\end{table}

\begin{table}[!t]
\centering
\resizebox{\columnwidth}{!}{
    \setlength\tabcolsep{1.5pt}
    \begin{tabular}{l|cccc}
    \toprule[1pt]
         & PSNR($\uparrow$) & SSIM($\uparrow$) & LPIPS($\downarrow$) & ABS($\downarrow$) \\ \hline
         Base model (\cref{table:abl_key} (d))            & 32.35 & 0.9245 & 0.1282 & 0.1416 \\  
         \hspace{3mm} + more depth planes                 & 32.33 & 0.9247 & 0.1299 & 0.1470 \\  
         \hspace{3mm} + DGPS (\cref{table:abl_key} (e))   & \textbf{33.09} & \textbf{0.9304} & \textbf{0.1230} & \textbf{0.1246} \\  
         
    \bottomrule[1pt]
    \end{tabular}
}
\vspace{-2mm}
\caption{
Efficiency of our depth-guided plane-sweeping (DGPS) compared to the baseline models using more depth planes.
}
\vspace{-2mm}
\label{table:abl_sweeping_add}
\end{table}

\begin{table}[!t]
\centering
\resizebox{\columnwidth}{!}{
    \setlength\tabcolsep{1.5pt}
    \begin{tabular}{l|cccc}
    \toprule[1pt]
         & PSNR ($\uparrow$) & SSIM ($\uparrow$) & LPIPS ($\downarrow$) & ABS ($\downarrow$) \\ \hline
         Ours w/ MVS depth ($d_{gt}^{mvs}$)      & 32.73 & 0.9277 & 0.1253 & 0.1216 \\ 
         Ours w/ Stereo depth ($d_{gt}$)   & \textbf{33.45} & \textbf{0.9336} & \textbf{0.1203} & \textbf{0.1056} \\ 
    \bottomrule[1pt]
    \end{tabular}
}
\vspace{-2mm}
\caption{
Effectiveness of using stereo depths as pseudo-ground truth for depth loss compared to using MVS depths.
}
\label{table:abl_depth}
\vspace{-4mm}
\end{table}

\section{Conclusion}
This paper proposes \MethodName{}, a novel generalizable view synthesis framework leveraging stereo-camera images for high-quality novel-view synthesis. 
Due to the ill-posedness, previous methods struggle with accurate geometry estimation, which leads to severe artifacts in novel-view synthesis.
Since the stereo matching provides vital information for accurate geometry reconstruction, our framework incorporates the stereo matching into NeRF-based generalizable view synthesis approach.
To this end, we introduce a stereo feature extractor, a depth-guided plane-sweeping, and a stereo depth loss.
We also present the StereoNVS dataset, the first stereo-camera image dataset for training and evaluating novel-view synthesis.
Our extensive experiments show that \MethodName{} is effective in generalizable novel-view synthesis, particularly in scenes with complex structures or textureless regions.

\vspace{-3.4mm}
\paragraph{Limitations and Future Work.} 
Our method is not free from limitations. In sparse view settings, our method produces blurry images and inaccurate geometry, issues also present in other methods.
This limitation arises from the insufficient information for novel viewpoints in the sparse view settings.
Our future work will involve additional geometric or generative prior, along with stereo prior, to compensate for the lack of information in such settings.

\vspace{-3.4mm}
\paragraph{Acknowledgement.}
We thank Woohyeok Kim and Hyeongmin Lee for their assistance in acquisition of the StereoNVS-Real dataset.
This work was supported by the NRF grant (No.2023R1A2C200494611, 2022R1A6A1A03052954, RS-2023-00211658) and IITP grant (No.2019-0-01906, Artificial Intelligence Graduate School Program (POSTECH)) funded by the Korea government (MSIT) and Samsung Electronics Co., Ltd.
{
    \small
    \bibliographystyle{ieeenat_fullname}
    \bibliography{main}
}





\end{document}


\clearpage
\setcounter{page}{1}
\maketitlesupplementary

\setcounter{section}{0}
\renewcommand\thesection{\Alph{section}}

\setcounter{table}{0}
\renewcommand\thetable{S\arabic{table}}

\setcounter{figure}{0}
\renewcommand\thefigure{S\arabic{figure}}

\setcounter{equation}{0}
\renewcommand\theequation{S\arabic{equation}}

\section{Overview}

In this supplementary document, we provide:
\begin{itemize}
    \item additional implementation details~(\cref{sec:sup_imp}),
    \item details on experimental settings~(\cref{sec:sup_exp}),
    \item additional analyses~(\cref{sec:sup_res_analysis}),
    \item details on the StereoNVS dataset~(\cref{sec:sup_data}), 
    \item additional quantitative results~(\cref{sec:sup_res_quan}), and
    \item additional qualitative results~(\cref{sec:sup_res_qual}).
\end{itemize}

\section{Implementation Details for \MethodName{}}
\label{sec:sup_imp}
\subsection{Stereo estimation network}
\label{sec:sup_imp_unimatch}
We employ UniMatch~\cite{xu2023unifying} as the stereo estimation network in our framework.
The stereo estimation network $S$ outputs multiple depths during its refinement stage:
\begin{equation}
\begin{aligned}
(d_{s,L}^{n,k},d_{s,R}^{n,k}) &= S(I_L^n,I_R^n), \\
\text{where} ~ k &\in \{1,\cdots,K\},
\end{aligned}
\end{equation}
where $K$ (set to 7) is the total number of outputs.
Throughout the paper, we use $(d_{s,L}^{n},d_{s,R}^{n})$ to indicate $(d_{s,L}^{n,K},d_{s,R}^{n,K})$ for simplicity, where $(d_{s,L}^{n,K},d_{s,R}^{n,K})$ are the final outputs.
These multiple outputs are supervised using the stereo depth loss, which will be explained in~\cref{sec:sup_loss}.
Refer to~\cite{xu2023unifying} for more details about the stereo estimation network.

\subsection{Stereo Feature Extractor}
\label{sec:sup_imp_stereo}
We provide additional details of the network architecture of the stereo feature extractor. 
To extract image features from input images, we modify the feature extractor from GeoNeRF~\cite{johari2022geonerf} as follows:
(1) We make the CNN encoder to project each of a stereo image pair $(I_L^n,I_R^n)$ to feature maps.
(2) We introduce the stereo attention module (SAM)~\cite{chu2022nafssr} to fuse feature maps from the input stereo image pair.
(3) We further extend SAM to aggregate stereo correlated features $(t_L^n,t_R^n)$.

We utilize the Feature Pyramid Network (FPN) for both the CNN encoder and the CNN decoder within the stereo feature extractor, as done in GeoNeRF.
The CNN decoder takes the fused feature maps from SAM as inputs and generates stereo image features with three different scales.
We denote the multi-scale stereo image features as $(f_L^{n,l},f_R^{n,l})$, where $l$ is a scale index such that $l\in\{0,1,2\}$.
$f_L^{n,2}$ has the same size as $I_L^n$, while $f_L^{n,0}$ and $f_L^{n,1}$ have resolutions of 1/4 and 1/2, respectively.
These multi-scale stereo image features will be utilized to build multi-scale cost volumes in subsequent feature volume construction~\cref{sec:sup_imp_feat}.

\subsection{Depth-guided Feature Volume Construction}
\label{sec:sup_imp_feat}
We provide a detailed description for the depth-guided feature volume construction~(Sec.~3.2).
We adopt the cascaded cost volume strategy~\cite{gu2020cascade} of GeoNeRF~\cite{johari2022geonerf} for constructing feature volumes.
The feature volumes provide occlusion-aware geometric information for conditioning the neural renderer.
To obtain fine and high-resolution feature volumes, cost volumes are built in a coarse-to-fine manner with three steps.
Then, we introduce our depth-guided plane-sweeping (DGPS) in the initial step of the cascaded cost volume strategy.
In the following, we explain the cascaded cost volume strategy first, followed by the incorporation of DGPS in this strategy.

The initial stage of the cascaded cost volume strategy begins by establishing the depth range using predefined near and far depths, denoted as $d_{near}$ and $d_{far}$, respectively.
Depth planes are then hypothesized by uniformly dividing the entire depth range by a predetermined number of depth planes, denoted as $M$.
The interval between consecutive depth planes is determined as $\frac{d_{far}-d_{near}}{M}$ during this initial stage.
On these depth planes, plane-sweeping aggregates the coarsest-scale multi-view features $(f_L^{n,0},f_R^{n,0})$ to build the initial-stage cost volumes.
Then, the MVS network produces the initial-stage feature volumes $(\phi_L^{n,0},\phi_R^{n,0})$ and the initial-stage depth maps $(d_{m,L}^{n,0},d_{m,R}^{n,0})$ from these initial-stage cost volumes.

Next, for the cost volume construction in the subsequent stages, we hypothesize depth planes around the depths obtained from the previous stage and perform plane-sweeping on these depth planes to aggregate finer-scale multi-view features $(f_L^{n,l},f_R^{n,l})$.
The plane interval of the finer-scale cost volume is reduced to $1/2$ compared to that of the previous stage.
The feature volumes and depth maps for each stage are obtained via the MVS network, similar to the initial stage.
We denote the resulting multi-scale feature volumes and depth maps as $(\phi_{L}^{n,l},\phi_{R}^{n,l})$ and $({d}_{m,L}^{n,l},{d}_{m,R}^{n,l})$, respectively.
We utilize individual UNet for the MVS network at each stage.
The number of hypothesis depth planes $M$ for each stage is set to 48, 32, and 8, respectively.
The resolution of $d_{m,L}^{n,l}$ are the same as $f_L^{n,l}$.

However, using only the cascaded cost volume strategy might not ensure accurate geometry estimation, especially when the initial depth estimation is not precise.
Inaccurate depth estimation may hinder the subsequent construction of feature volumes around the real geometry.
Since the prediction of NeRF relies on these feature volumes, the neural renderer predicts inaccurate geometry when feature volumes are not constructed around the real geometry. 
To tackle this issue, we introduce DGPS in the first stage of this strategy.
Unlike the previous approach that hypothesizes depth planes across the entire depth range, DGPS hypothesizes depth planes around the stereo depth $(d_{s,L}^{n},d_{s,R}^{n})$ obtained from the stereo estimation network.
Leveraging the reliable stereo depth, DGPS ensures feature volume construction around the geometry, as explained in~Sec.~5.3.

\subsection{Stereo Depth Loss}
\label{sec:sup_loss}
We provide detailed explanation of Eq.~3 in the main paper:
$\mathcal{L}_d^{s}$ and $\mathcal{L}_d^{m}$ are computed in a pixel-wise manner, and $\mathcal{L}_d^{r}$ is computed in a ray-wise manner.
Specifically, $\mathcal{L}_d^{s}$ and $\mathcal{L}_d^{m}$ are defined as:
\begin{eqnarray}
\hspace{-0.5cm} \mathcal{L}_d^{s} \hspace{-0.3cm} &=& \hspace{-0.3cm} \sum_{n=1}^N \sum_{k=1}^K \gamma^{K-k} \left\{ g(d_{s,L}^{n,k}, d_{gt,L}^{n}) + g(d_{s,R}^{n,k}, d_{gt,R}^{n}) \right\}, \\
\hspace{-0.5cm} \mathcal{L}_d^{m} \hspace{-0.3cm} &=& \hspace{-0.3cm} \sum_{n=1}^N \sum_{l=0}^2 2^{-l} \left\{ g(d_{m,L}^{n,l}, d_{gt,L}^{n,l}) + g(d_{m,R}^{n,l}, d_{gt,R}^{n,l})\right\},
\end{eqnarray}
where $g(d,d_{gt})$ represents the distance between the estimated depth maps $d$ and the pseudo-ground-truth depth maps $d_{gt}$.
$K$ is the total number of depth predictions and $\gamma$ (set to 0.9) is the weight to give higher weights for later depth prediction~(\cref{sec:sup_imp_unimatch}).
$d_{gt,L}^n$ and $d_{gt,R}^n$ are pseudo-ground-truth depth maps of $I_L^n$ and $I_R^n$, respectively.
$d_{gt,L}^{n,l}$ and $d_{gt,R}^{n,l}$ are the downsampled versions of $d_{gt,L}^n$ and $d_{gt,R}^n$ according to the scale level $l$, respectively.

The distance function $g$ is defined as:
\begin{equation}
\begin{aligned}
g(d,d_{gt}) &= \sum_{ij} m(i,j) \left\| \frac{1}{d(i,j)} - \frac{1}{d_{gt}(i,j)} \right\|,
\end{aligned}
\end{equation}
where $(i,j)$ is a pixel coordinate, and $m(i,j)$ is a mask for the pixel $(i,j)$.
The distance function computes the difference between the inverses of depth values.
The mask $m(i,j)$ is introduced to account for errors in the pseudo-ground-truth depth map.
Specifically, we define $m(i,j)=1$ if $d_{gt}(i,j)$ falls within the boundaries defined by the near and far depth range from COLMAP~\cite{schonberger2016structure}, and $m(i,j)=0$ otherwise.
For the norm $\|\cdot\|$, we use a smooth $L_1$ norm.

Among the loss terms of the stereo depth loss, $\mathcal{L}_d^{r}$ is defined as follow:
\begin{equation}
\begin{aligned}
\label{equ:xxx}
\mathcal{L}_d^{r} = \sum_{\textbf{\text{r}} \in \mathcal{R}} m(\textbf{\text{r}}) \left\| \frac{1}{d(\textbf{\text{r}})} - \frac{1}{d_{gt}(\textbf{\text{r}})} \right\|,
\end{aligned}
\end{equation}
where $\mathcal{R}$ denotes set of rays in each training batch, and $m(\textbf{\text{r}})$ is a mask for ray $\textbf{\text{r}}$.
$m(\textbf{\text{r}})$ is defined in the same way as $m(i,j)$.

\subsection{Training Details for \MethodName{}}
We set the balancing weights $\lambda_{d}^{self}$ and $\lambda_{d}^{stereo}$ in Eq.~2 as 0.1 and 1.0, respectively.
For the stereo depth loss $\mathcal{L}_{d}^{stereo}$ in Eq.~3, we set $\lambda_{1}$, $\lambda_{2}$, and $\lambda_{3}$ as 0.1, 0.1, and 1.0, respectively.
The image resolution for training and evaluation is $512 \times 256$ for both the StereoNVS-Real and the StereoNVS-Synthetic datasets.
We use UniMatch~\cite{xu2023unifying} pre-trained on several mixed public datasets, which is used for all the mentioned pre-trained stereo estimation networks in our framework.
Specifically, we employ the configuration of `GMStereo-scale2-regrefine3-resumeflowthings-mixdata` from the official UniMatch website (\href{https://github.com/autonomousvision/unimatch}{https://github.com/autonomousvision/unimatch}).
We use CasMVSNet~\cite{gu2020cascade} pre-trained on BlendedMVS~\cite{yao2020blendedmvs} for our MVS network, as done in GeoNeRF~\cite{johari2022geonerf}.
The training takes 5 days using one RTX3090 GPU.

\subsection{Computing Depth from Disparity}
We utilize the stereo estimation network to obtain pseudo-ground-truth depth $d_{gt}$ and stereo depths $d_{s}$.
Specifically, we employ UniMatch~\cite{xu2023unifying} for the stereo estimation network, which estimates disparities from input stereo-image pairs.
If the focal length and the baseline length between the stereo image pair are known, depths can be computed from the estimated disparities as follows:
\begin{equation}
\begin{aligned}
d = \frac{b \times f}{\text{disp}},
\end{aligned}
\end{equation}
where $d$, $b$, $f$, and $\text{disp}$ denote depth, baseline length, focal length, and disparity, respectively.
We use the pre-computed values for the baseline length and the focal length~(\cref{sec:sup_data}).


\begin{figure*}[!t]
\centering
\begin{tabular}{@{}c}
\includegraphics[width=\linewidth]{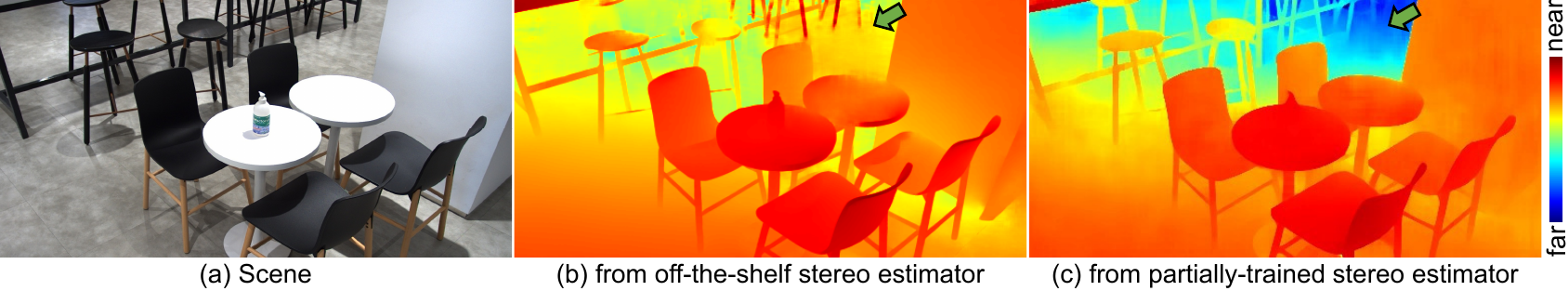} \\
\end{tabular}
\vspace{-2mm}
\caption{
Improved depth estimation accuracy thanks to our partial training scheme for the stereo estimation network.
Without partial training scheme, the off-the-shelf stereo estimation network~\cite{xu2023unifying} predicts depths with an incorrect scale.
}
\label{fig:sup_partial_train}
\end{figure*}

\section{Experimental Settings}
\label{sec:sup_exp}
\subsection{Training Details for Baselines}
We train the baseline methods: SRF~\cite{chibane2021stereo}, IBRNet~\cite{wang2021ibrnet}, GNT~\cite{varma2022attention}, GeoNeRF~\cite{johari2022geonerf}, and NeuRay~\cite{liu2022neural} on the training set of the StereoNVS-Real dataset.
We use 512 rays for the training batch except for GNT.
All the models are trained for 250K iterations.

\paragraph{SRF.}
For SRF~\cite{chibane2021stereo}, we use official PyTorch implementation from \href{https://github.com/jchibane/srf}{https://github.com/jchibane/srf}. 
For training, we use the Adam optimizer with learning rate of 0.0005 and the exponential decay strategy for learning rate scheduling.

\paragraph{IBRNet.}
For IBRNet~\cite{wang2021ibrnet}, we use official PyTorch implementation from \href{https://github.com/googleinters/IBRNet}{https://github.com/googleinters/IBRNet}. 
For training, we use the Adam optimizer with learning rates of 0.001 and 0.0005 for feature extraction and rendering network, respectively. 
we use the exponential decay strategy for learning rate scheduling.

\paragraph{GNT.}
For GNT~\cite{varma2022attention}, we use official PyTorch implementation from \href{https://github.com/VITA-Group/GNT}{https://github.com/VITA-Group/GNT}. 
For training, we use the we use the Adam optimizer with learning rates of 0.001 and 0.0005 for feature extraction and rendering network.
we use the exponential decay strategy for learning rate scheduling. 
Since GNT employs large transformer backbone, we use 2048 rays for its training batch.

\paragraph{GeoNeRF.}
For GeoNeRF~\cite{johari2022geonerf}, we use official PyTorch implementation from \href{https://github.com/idiap/GeoNeRF}{https://github.com/idiap/GeoNeRF}. 
For training, we use the Adam optimizer with learning rates of 0.0005 and the cosine-annealing scheduling.
We use CasMVSNet~\cite{gu2020cascade} pre-trained on BlendedMVS~\cite{yao2020blendedmvs} for the MVS network of GeoNeRF.

\paragraph{NeuRay.}
For NeuRay~\cite{liu2022neural}, we use official PyTorch implementation from \href{https://github.com/liuyuan-pal/NeuRay}{https://github.com/liuyuan-pal/NeuRay}.
For training, we use the Adam optimizer with learning rates of 0.0002 and the exponential decay strategy for learning rate scheduling.

\subsection{Experimental Details for Analysis}
We provide additional details for experiments in Sec.~5.3.2 and Sec.~5.3.3 in the main paper.
\subsubsection{Effectiveness of Depth-Guided Plane-Sweeping}
Tab.~4 in the main paper shows the results using more depth planes for cascaded cost volume construction. 
As DGPS is applied in the first stage of the cascaded cost volume approach, we utilize 96 depth planes in the first stage for the additional model.
Note that we use 48 depth planes in the first stage for all the other models.

\subsubsection{Benefit of Stereo Estimation in Depth Loss}
Tab.~5 in the main paper shows the effectiveness of utilizing stereo estimation networks for depth supervision.
To obtain $d_{gt}^{mvs}$, we use UniMVSNet~\cite{peng2022rethinking}, one of the state-of-the-art MVS network, which is trained on the DTU dataset~\cite{jensen2014large} and the BlendedMVS dataset~\cite{yao2020blendedmvs}.

\begin{table}[!t]
\centering
\resizebox{\columnwidth}{!}{
    \setlength\tabcolsep{1.5pt}
    \begin{tabular}{l|cccc}
    \toprule[1pt]
         & PSNR (\uparrow) & SSIM (\uparrow) & LPIPS (\downarrow) & ABS (\downarrow) \\ \hline
         Model w/o partial training scheme     & 32.52 & 0.9252 & 0.1299 & 0.1287 \\ 
         Our final model                    & 33.45 & 0.9336 & 0.1203 & 0.1056 \\ 
    \bottomrule[1pt]
    \end{tabular}
}
\caption{
Effectiveness of our partial training scheme for robust depth estimation of the stereo estimation network.
}
\label{table:freeze_unimatch}
\vspace{-2mm}
\end{table}

\section{Additional Analyses on \MethodName{}}
\label{sec:sup_res_analysis}
\subsection{Robustness to Stereo Depth Error}
In our experimental setup, we utilize pseudo-ground truth stereo depth maps estimated from an off-the-shelf model~\cite{xu2023unifying} for NeuRay~\cite{liu2022neural}, GeoNeRF$_{+D}$~\cite{johari2022geonerf}, and our proposed method. 
Since NeuRay and GeoNeRF$_{+D}$ rely on depth without considering potential errors, they are not robust to depth inaccuracies. As shown in Tab.~1 in the main paper, these methods report significantly degraded results in real-world scenes where obtaining error-free depth proves challenging. 
However, our approach circumvents this issue by incorporating partial training of the stereo estimation network within the stereo feature extractor, leveraging multi-view supervision (Sec.~3.4 in the main paper).
This training scheme ensures robust depth estimation of the stereo estimation network, subsequently utilized in our DGPS, resulting in consistent performance regardless of the dataset.
In the following discussion, we present additional experiments to further demonstrate the robustness of our method to depth errors.

\cref{fig:sup_partial_train}~(b) shows such estimation errors where an off-the-shelf depth estimation model inaccurately predicts depths with incorrect scales.
For example, the depth of distant regions is often inaccurately estimated to be closer to the camera.
This erroneous depth estimation negatively affects the performance of GeoNeRF$_{+D}$ and NeuRay, particularly in real-world scenes (Tab.~1 in the main paper). However, our proposed solution, explained in Sec.~3.4 in the main paper, effectively mitigates this issue by incorporating partial training of the stereo estimation network.

To demonstrate the effectiveness of our partial training scheme of the stereo estimation network, we conduct an additional experiment.
Specifically, we train an additional model, which is our final model with frozen parameters of the stereo estimation network.
We then compare the additional model with our final model based on both image and shape qualities.
\cref{table:freeze_unimatch} presents the results, indicating a slight decline in performance for the additional model compared to our final model employing the partial training scheme.

This slight performance decline can be attributed to two main factors.
Firstly, the use of inaccurate depth in DGPS results in the misplacement of feature volumes, deviating from the actual geometry. 
This misalignment, in turn, leads to inaccurate geometry estimation by the neural renderer, negatively impacting view synthesis performance. 
Secondly, as we use rendered depth as ground truth for $L_d^{self}$, the depth error introduced by the neural renderer can destabilize our framework's training process.
In contrast, our proposed partial training scheme enhances the stereo estimation network's ability to estimate depth with precise scale (\cref{fig:sup_partial_train}~(c)).
It is because this training scheme makes our stereo estimation network more robust to differences in dataset characteristics.
Therefore, DGPS with more accurate depth ensures the construction of feature volumes around real geometry, ultimately leading to improved geometry estimation by the neural renderer. 
This enhanced accuracy contributes to stable training, leading to better performance.

\begin{table}[!t]
\centering
\resizebox{\columnwidth}{!}{
    \setlength\tabcolsep{1.5pt}
    \begin{tabular}{c|l|ccc}
    \toprule[1pt]
         Crop Ratio     & Method            & PSNR (\uparrow)   & SSIM (\uparrow)   & LPIPS (\downarrow) \\ \hline
          1             & GeoNeRF$^{(6)}$   & \textbf{28.77}    & \textbf{0.907}    & \textbf{0.139} \\ 
                        & Ours              & 28.44             & 0.900             & 0.140 \\ \hline
         1/2            & GeoNeRF$^{(6)}$   & \textbf{28.87}    & 0.923             & 0.094 \\ 
                        & Ours              & 28.84             & \textbf{0.924}    & \textbf{0.085} \\ \hline
         1/3            & GeoNeRF$^{(6)}$   & 29.09             & 0.926             & 0.093 \\ 
                        & Ours              & \textbf{29.23}    & \textbf{0.931}    & \textbf{0.080} \\ 
    \bottomrule[1pt]
    \end{tabular}
}
\caption{
Comparison stereo-camera setting against monocular-view setting with center-cropping strategy.
}
\label{table:comp_mono}
\vspace{-3mm}
\end{table}

\subsection{Comparison Between Monocular-Camera and Stereo-Camera Settings}
We conduct further evaluation of the effectiveness of the stereo-camera setting by comparing our method with a baseline model~\cite{johari2022geonerf}, which is trained on six views (i.e., six images) in a monocular setting. Tab.~2 in the main paper already demonstrates that our proposed method, trained on three stereo-camera pairs (i.e., six images), outperforms the baseline method trained on three views (i.e., three images) in the monocular setting. To provide a more comprehensive comparison, we present an additional experiment comparing the monocular-camera and stereo-camera settings using the same number of images (i.e., six images).

In this experiment, we compare our method against the baseline method, GeoNeRF~\cite{johari2022geonerf}, specifically a variant denoted as GeoNeRF$^{(6)}$, which utilizes six monocular images in the inference. 
These six monocular images are obtained by selecting the left-side images from six stereo-camera image pairs. 
Note that leveraging six monocular views significantly broadens the observation of spatial views. 
To mitigate the impact of unobserved spatial regions in the evaluation, we compute errors by comparing center-cropped regions of both synthesized and ground-truth images.

As shown in \cref{table:comp_mono}, leveraging additional spatial views, GeoNeRF$^{(6)}$ achieves better results than our method. 
However, employing a 1/2 center-cropping strategy in the comparison, our method shows comparable or slightly better results than GeoNeRF$^{(6)}$. 
In addition, employing a 1/3 center-cropping strategy, our method surpasses GeoNeRF$^{(6)}$. These results underscore the effectiveness of our method, particularly in the observed region.

\subsection{Choice of Stereo Estimation Network}

In our framework, we adopt UniMatch~\cite{xu2023unifying} as the stereo estimation network due to its generalization capability. 
This capability is attributed to its training on a large-scale stereo-image dataset, which endows UniMatch with the ability to generalize well across various datasets. Consequently, our framework benefits from this high generalization capacity, leading to superior performance. 
To validate the relationship between the performance of our framework and the generalization capability of the stereo estimation network, we conduct an experiment comparing our final model with another model employing a different stereo estimation network.

This alternative model utilizes a stereo estimation network that has the identical network architecture to the stereo estimation network in our final model but is trained on the SceneFlow dataset~\cite{MIFDB16}.
Specifically, we utilize the ``GMStereo-scale2-regrefine3-sceneflow'' configuration from the official UniMatch website (\href{https://github.com/autonomousvision/unimatch}{https://github.com/autonomousvision/unimatch}) for the stereo estimation network.
Since the SceneFlow dataset is notably smaller in size compared to the large-scale dataset used to train our final model, this alternative model lacks the generalization ability demonstrated by our final model.

We compare the performances of our final model and this alternative model on the StereoNVS-Synthetic dataset.
As shown in~\cref{table:stereo_choice}, our final model outperforms this alternative model across various metrics. This result underscores the correlation between the generalization ability of the stereo estimation network and the performance of our framework. Furthermore, we anticipate that incorporating a more refined stereo estimation network into our framework has the potential to further enhance performance.

\begin{table}[!t]
\centering
\resizebox{\columnwidth}{!}{
    \setlength\tabcolsep{1.0pt}
    \begin{tabular}{l|cccc}
    \toprule[1pt]
         & PSNR (\uparrow) & SSIM (\uparrow) & LPIPS (\downarrow) & ABS (\downarrow) \\ \hline
         Model w/ small-dataset-trained stereo network     & 31.73 & 0.9026 & 0.1637 & 0.2207 \\ 
         Our final model     & \textbf{33.45} & \textbf{0.9336} & \textbf{0.1203} & \textbf{0.1056} \\ 
    \bottomrule[1pt]
    \end{tabular}
}
\vspace{-2mm}
\caption{
Experiment on the choice of the stereo estimation network in our framework.
}
\label{table:stereo_choice}
\vspace{-3mm}
\end{table}

\begin{figure*}[!t]
\centering
\begin{tabular}{@{}c}
\includegraphics[width=\linewidth]{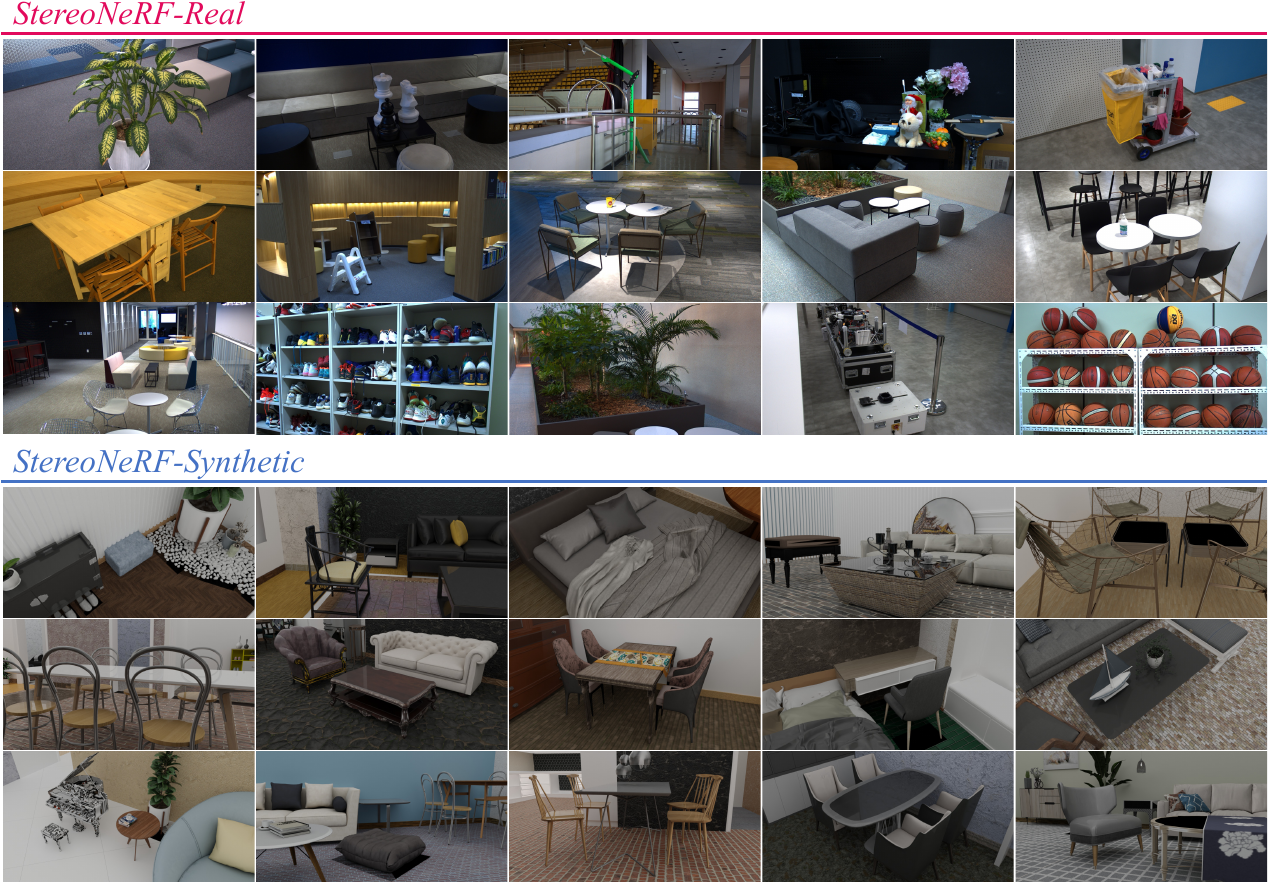} \\
\end{tabular}
\caption{
The StereoNVS datasets.
The StereoNVS dataset presents multi-view stereo-pair images for both real-world and synthetic scenes.
These images are part of the entire dataset.
}
\label{fig:dataset}
\end{figure*}

\section{StereoNVS dataset}
\label{sec:sup_data}

\cref{fig:dataset} shows example scenes in the StereoNVS-Real and the StereoNVS-Synthetic datasets.

For the StereoNVS-Real, the baseline length is around 8.8cm and the focal length is around 1568 pixels.
These values vary slightly for each scene due to the scale ambiguity induced by the structure-from-motion approach of COLMAP for camera pose acquisition.
When capturing the real-world scenes, we manually fixed the stereo camera's focal length, exposure time, gain, gamma correction, and white balance for each scene.
In addition, since acquired images have a stereoscopic constraint with fixed camera positions and orientations, we employed the ``RigBundleAdjuster'' configuration in COLMAP~\cite{schonberger2016structure} to ensure the stereoscopic constraint during the camera pose acquisition.

For the StereoNVS-Synthetic, the baseline length and the focal length are 8cm and 711 pixels, respectively.
We randomly sampled 50 synthetic scenes from the 3D-Front dataset~\cite{fu20213d}.
For each scene, we generated 200 viewpoints of stereo image pairs, which are constrained to look at a specific object like furniture.
Then, we filtered camera viewpoints placed too close to the objects in the scene.

\section{Additional Quantitative Results}
\label{sec:sup_res_quan}

\begin{table}[!t]
\centering
\begin{adjustbox}{width=0.3\textwidth}
\resizebox{\columnwidth}{!}{
    \setlength\tabcolsep{1.5pt}
    \begin{tabular}{c|ccc}
    \hline
        Method & ABS (\downarrow) & ARE (\downarrow) & RMSE (\downarrow)\\ \hline
        SRF            & 0.8125 & 0.4673 & 1.1347  \\
        IBRNet         & 0.2628 & 0.1408 & 0.5137 \\
        GeoNeRF        & 0.1577 & 0.0933 & 0.3631 \\
        GNT            & 0.4512 & 0.1945 & 0.6593 \\  \hline
        GeoNeRF$_{+D}$ & 0.0782 & 0.0576 & 0.2301 \\  
        NeuRay         & 0.1571 & 0.0852 & 0.3636 \\ \hline
        Ours           & 0.1056 & 0.0725 & 0.2656 \\ \hline
    \end{tabular}
}
\end{adjustbox}
\caption{Depth quality assessment using additional depth metrics. We compare our method against the baseline methods~\cite{chibane2021stereo,wang2021ibrnet,johari2022geonerf,varma2022attention,liu2022neural}. Our method shows comparable depth quality to GeoNeRF$_{+D}$, although GeoNeRF$_{+D}$ explicitly uses depth maps in the inference. 
}
\label{table:synthetic_depth_more_metric}
\end{table}

\subsection{Depth Quality Assessment}
We assess the depth quality using additional depth metrics such as absolute error (ABS), absolute relative error (ARE), and root mean square error (RMSE).
~\cref{table:synthetic_depth_more_metric} presents depth quality evaluation for our method and all the baseline methods~\cite{wang2021ibrnet,johari2022geonerf,varma2022attention,liu2022neural} on the StereoNVS-Synthetic dataset.
The result reveals similar trends to those using ABS in the Tab.~1 in the main paper.
Our method shows comparable depth quality to GeoNeRF$_{+D}$, although GeoNeRF$_{+D}$ explicitly uses depth maps in the inference.

\subsection{Evaluation on the BlendedMVS Dataset}
We evaluate our model on the BlendedMVS dataset~\cite{yao2020blendedmvs}.
Since BlendedMVS do not provide stereo-camera images, we were not able to evaluate our model directly on BlendedMVS, which is released on the official website (\href{https://github.com/YoYo000/BlendedMVS}{https://github.com/YoYo000/BlendedMVS}).
Nonetheless, while BlendedMVS does not directly provide stereo-camera images, it still provides 3D meshes.
Thus, we conducted an additional evaluation by rendering stereoscopic images from the meshes.
For this evaluation, we classify large scenes based on the information available on the website (\href{https://github.com/kwea123/BlendedMVS_scenes}{https://github.com/kwea123/BlendedMVS\_scenes}). Then, we sample several scenes from these categorized scenes.
The list of the names for the sampled scenes is as follows:
\begin{itemize}
    \item `5aa515e613d42d091d29d300`,
    \item `5bf18642c50e6f7f8bdbd492`,
    \item `5af02e904c8216544b4ab5a2`,
    \item `5b69cc0cb44b61786eb959bf`, 
    \item `5bfc9d5aec61ca1dd69132a2`,
    \item `5b08286b2775267d5b0634ba`,
    \item `5ba75d79d76ffa2c86cf2f05`,
    \item `58eaf1513353456af3a1682a`, and
    \item `5af28cea59bc705737003253`.
\end{itemize}

~\cref{table:blendedmvs_comp} shows that our method outperforms other baseline methods~\cite{chibane2021stereo,wang2021ibrnet,johari2022geonerf,varma2022attention,liu2022neural} on the BlendedMVS dataset.
Given the complex structures in scenes from BlendedMVS, these results highlight the effectiveness of our method, particularly in such settings.
In addition, SRF~\cite{chibane2021stereo} and GNT~\cite{varma2022attention} demonstrate significantly degraded performances on BlendedMVS due to their limited generalization capabilities.

\begin{table}[!t]
\centering
\resizebox{\columnwidth}{!}{
    \setlength\tabcolsep{1.5pt}
    \begin{tabular}{c|cccccc}
    \hline
        Method & PSNR(\uparrow) & SSIM(\uparrow) & LPIPS(\downarrow) & ABS (\downarrow) & ARE (\downarrow) & RMSE (\downarrow)\\ \hline
        SRF            & 13.93 & 0.2251 & 0.6866   & 39.74 & 0.5952 & 50.96 \\
        IBRNet         & 21.68 & 0.6758 & 0.3191   & 7.726 & 0.1539 & 15.86 \\
        GeoNeRF        & 23.31 & 0.7635 & 0.2426   & 4.552 & 0.0769 & 11.45 \\
        GNT            & 11.58 & 0.2459 & 0.7675   & 23.31 & 0.4636 & 32.04 \\  \hline
        GeoNeRF$_{+D}$ & 23.67 & 0.7390 & 0.2618   & 3.434 & \textbf{0.0473} & \textbf{8.011} \\  
        NeuRay         & 23.29 & 0.7271 & 0.2700   & 3.358 & 0.0634 & 8.861 \\ \hline
        Ours           & \textbf{24.08} & \textbf{0.7842} & \textbf{0.2301}   & \textbf{3.275} & 0.0572 & 8.452 \\ \hline
    \end{tabular}
}
\caption{Evaluation on the BlendedMVS dataset~\cite{yao2020blendedmvs}. We quantitatively compare our method against the baseline methods~\cite{chibane2021stereo,wang2021ibrnet,johari2022geonerf,varma2022attention,liu2022neural}.
}
\label{table:blendedmvs_comp}
\vspace{-2mm}
\end{table}

\begin{table*}[!t]
\centering
\begin{tabular}{@{}c}
    \resizebox{\textwidth}{!}{

    \begin{tabular}{l|c|c|c|c|c|c|c|c}
    \toprule[1pt]
         (mean,var)         & Scene \#3    & Scene \#11   & Scene \#13   & Scene \#25   & Scene \#26   & Scene \#30   & Scene \#38   & Scene \#40 \\ \hline
         IBRNet             & (26.58, 1.5) & (27.19, 1.4) & (21.32, 5.6) & (28.52, 3.5) & (28.41,2.8) & (23.32, 4.7) & (27.50, 1.6) & (26.41, 1.7) \\ \hline
         GeoNeRF            & (28.40, 2.7) & (28.81, 1.8) & (\textbf{22.88}, 4.5) & (30.46, 2.6) & (30.23, 3.2) & (24.93, 4.6) & (30.05, 2.2) & (28.60, 2.6) \\ \hline
         GNT                & (26.00, 1.2) & (27.17, 1.5) & (21.38, 3.5) & (28.86, 2.9) & (28.61, 1.7) & (23.30, 4.5) & (27.51, 1.1) & (25.96, 2.6) \\ \hline
         GeoNeRF$_{+D}$     & (27.12, 2.4) & (26.50, 1.0) & (21.54, 3.2) & (28.26, 3.4) & (29.00, 1.9) & (23.32, 2.8) & (27.64, 1.1) & (27.59, 0.8) \\ \hline
         NeuRay             & (27.73, 2.2) & (27.61, 1.8) & (21.54, 3.2) & (28.26, 3.4) & (29.00, 1.9) & (23.32, 2.8) & (27.64, 1.1) & (27.59, 0.8) \\ \hline
         Ours               & (\textbf{28.87}, 2.6) & (\textbf{29.35}, 2.2) & (22.85, 4.5) & (\textbf{30.82}, 2.4) & (\textbf{30.76}, 3.8) & (\textbf{25.37}, 5.0) & (\textbf{30.50}, 2.2) & (\textbf{29.33}, 2.4) \\ 
    \bottomrule[1pt]
    \end{tabular}
    }\\
    {(a) PSNR}\\\\    

    \resizebox{\textwidth}{!}{
    \setlength\tabcolsep{1.5pt}
    \begin{tabular}{l|c|c|c|c|c|c|c|c}
    \toprule[1pt]
         (mean,var)         & Scene \#3         & Scene \#11        & Scene \#13        & Scene \#25        & Scene \#26        & Scene \#30        & Scene \#38        & Scene \#40        \\ \hline
         IBRNet             & (0.8133, 0.0007)  & (0.8953, 0.0004)  & (0.8386, 0.0038)  & (0.8729, 0.0010)  & (0.8239, 0.0009)  & (0.8491, 0.0018)  & (0.7850, 0.0007)  & (0.8675, 0.0005)   \\ \hline
         GeoNeRF            & (0.8669, 0.0005)  & (0.9210, 0.0002)  & (\textbf{0.8704}, 0.0019)  & (0.9173, 0.0003)  & (0.8801, 0.0005)  & (0.8897, 0.0013)  & (0.8967, 0.0011)  & (0.9034, 0.0005)   \\ \hline
         GNT                & (0.8077, 0.0005)  & (0.8977, 0.0002)  & (0.8516, 0.0018)  & (0.8802, 0.0006)  & (0.8363, 0.0007)  & (0.8518, 0.0016)  & (0.7999, 0.0009)  & (0.8205, 0.0009)   \\ \hline
         GeoNeRF$_{+D}$     & (0.8283, 0.0006)  & (0.8728, 0.0003)  & (0.7371, 0.0023)  & (0.7944, 0.0015)  & (0.8042, 0.0013)  & (0.7946, 0.0019)  & (0.8533, 0.0008)  & (0.8761, 0.0004)   \\ \hline
         NeuRay             & (0.8515, 0.0006)  & (0.8993, 0.0003)  & (0.8369, 0.0019)  & (0.8736, 0.0011)  & (0.8534, 0.0004)  & (0.8570, 0.0011)  & (0.7860, 0.0007)  & (0.8864, 0.0002)   \\ \hline
         Ours               & (\textbf{0.8726}, 0.0005) & (\textbf{0.9254}, 0.0002) & (0.8699, 0.0021) & (\textbf{0.9240}, 0.0002) & (\textbf{0.8917}, 0.0006) & (\textbf{0.8993}, 0.0010) & (\textbf{0.9056}, 0.0009) & (\textbf{0.9135}, 0.0004) \\ 
    \bottomrule[1pt]
    \end{tabular}
    }\\
    {(b) SSIM}\\\\

    \resizebox{\textwidth}{!}{
    \setlength\tabcolsep{1.5pt}
    \begin{tabular}{l|c|c|c|c|c|c|c|c}
    \toprule[1pt]
         (mean,var)         & Scene \#3         & Scene \#11        & Scene \#13        & Scene \#25        & Scene \#26        & Scene \#30        & Scene \#38        & Scene \#40        \\ \hline
         IBRNet             & (0.2812, 0.0007)  & (0.1811, 0.0008)  & (0.1744, 0.0020)  & (0.1729, 0.0009)  & (0.2400, 0.0013)  & (0.1889, 0.0011)  & (0.2144, 0.0009)  & (0.2349, 0.0009)   \\ \hline
         GeoNeRF            & (\textbf{0.1931}, 0.0010)  & (0.1221, 0.0001)  & (0.1343, 0.0009)  & (0.1139, 0.0003)  & (0.1658, 0.0009)  & (0.1513, 0.0011)  & (0.1160, 0.0003)  & (0.1832, 0.0006)   \\ \hline
         GNT                & (0.3189, 0.0007)  & (0.2026, 0.0008)  & (0.1741, 0.0012)  & (0.1881, 0.0008)  & (0.2672, 0.0011)  & (0.2043, 0.0014)  & (0.2262, 0.0012)  & (0.2643, 0.0014)   \\ \hline
         GeoNeRF$_{+D}$     & (0.2305, 0.0014)  & (0.1626, 0.0002)  & (0.2691, 0.0013)  & (0.1988, 0.0009)  & (0.2100, 0.0006)  & (0.2324, 0.0017)  & (0.1502, 0.0004)  & (0.2079, 0.0003)   \\ \hline
         NeuRay             & (0.2389, 0.0011)  & (0.1581, 0.0006)  & (0.1770, 0.0012)  & (0.1675, 0.0010)  & (0.1914, 0.0006)  & (0.1771, 0.0009)  & (0.2154, 0.0008)  & (0.1868, 0.0006)   \\ \hline
         Ours               & (0.1963, 0.0012) & (\textbf{0.1173}, 0.0002) & (\textbf{0.1329}, 0.0012) & (\textbf{0.1081}, 0.0003) & (\textbf{0.1482}, 0.0007) & (\textbf{0.1428}, 0.0007) & (\textbf{0.1094}, 0.0004) & (\textbf{0.1749}, 0.0005) \\ 
    \bottomrule[1pt]
    \end{tabular}
    }\\
    {(c) LPIPS}\\\\
\end{tabular}
\caption{
Evaluation using per-scene mean and variance of PSNR, SSIM, and LPIPS on the StereoNVS-Real test set.
}
\label{table:per_scene_metrics}
\end{table*}

\subsection{Evaluation Using Per-scene Metrics}
\cref{table:per_scene_metrics} presents per-scene evaluaion using mean and variance of PSNR, SSIM~\cite{wang2004image}, and LPIPS~\cite{zhang2018unreasonable} on the StereoNVS-Real test set. Our method mostly shows better performance than the baseline methods~\cite{wang2021ibrnet,johari2022geonerf,varma2022attention,liu2022neural}.

\section{Additional Qualitative Results}
\label{sec:sup_res_qual}
We present an additional qualitative comparison of \MethodName{} with other baseline methods: IBRNet~\cite{wang2021ibrnet}, GNT~\cite{varma2022attention}, GeoNeRF~\cite{johari2022geonerf}, and NeuRay~\cite{liu2022neural}.

\begin{figure*}[!t]
\centering
\begin{tabular}{@{}c}
\includegraphics[width=.925\linewidth]{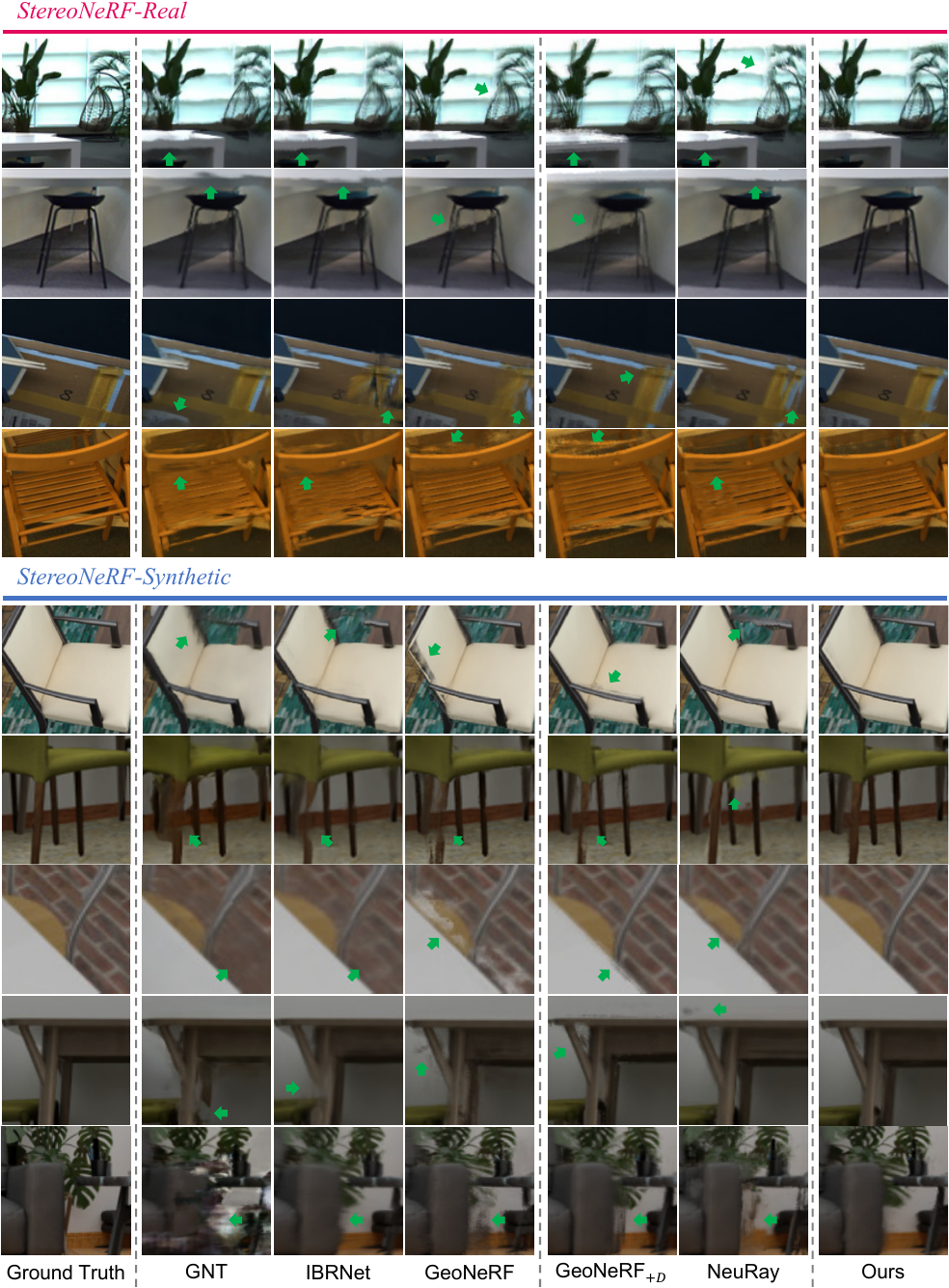} \\
\end{tabular}
\vspace{-3mm}
\caption{
Qualitative comparison of novel-view synthesis on the StereoNVS-Real and StereoNVS-Synthetic datasets.
All models are trained using stereo images.
Our method surpasses the baseline methods~\cite{wang2021ibrnet,varma2022attention,johari2022geonerf,liu2022neural}, especially for thin structures and textureless regions.
}
\vspace{-3mm}
\label{fig:supple_qual_main_path_patch}
\end{figure*}

\cref{fig:supple_qual_main} presents novel-view rendering results with whole images and depths for the StereoNVS-Real dataset.
Our method shows the best image quality, along with the accurately estimated depth maps.
While GeoNeRF$_{+D}$ seems to produce depth results comparable to our method, their synthesized images show artifacts, especially for the object boundary regions.
It is because they utilize the pseudo-ground-truth depths in its inference without considering the error.
Additionally, we present the rendering results of diverse scenes in the both StereoNVS-Real and StereoNVS-Synthetic datasets.
\cref{fig:supple_qual_main_path_patch} shows zoomed-in patches, where other baseline methods synthesize degenerated images, especially for thin structures and textureless regions.

\begin{figure*}[!t]
\centering
\begin{tabular}{@{}c}
\includegraphics[width=0.85\linewidth]{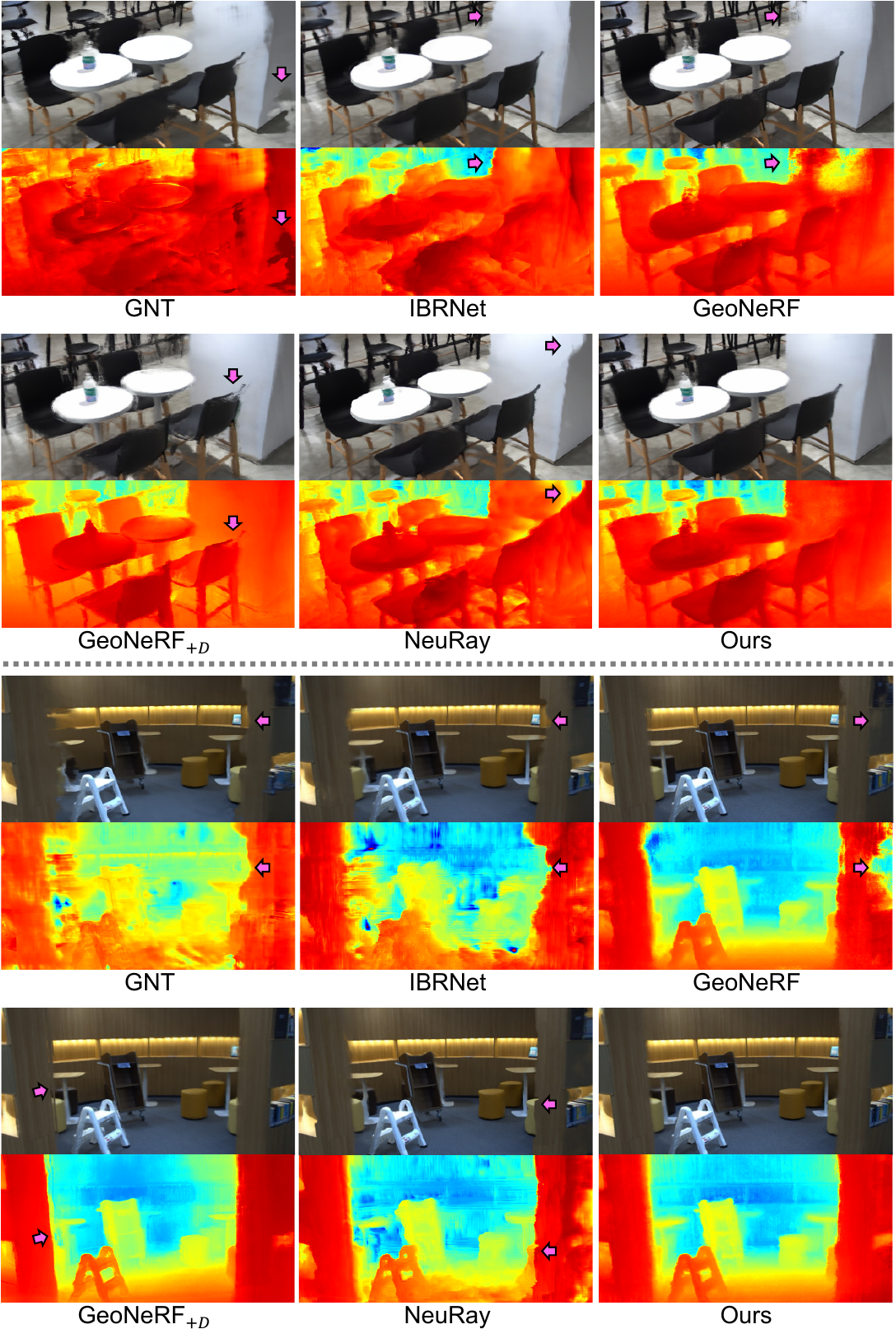} \\
\end{tabular}
\vspace{-3mm}
\caption{
Qualitative comparison of novel-view synthesis on the StereoNVS-Real dataset.
All models are trained using stereo images.
Our method outperforms the baseline methods~\cite{wang2021ibrnet,varma2022attention,johari2022geonerf,liu2022neural} for both image and depth qualities.
}
\vspace{-3mm}
\label{fig:supple_qual_main}
\end{figure*}

\begin{figure*}[!t]
\ContinuedFloat 
\centering
\begin{tabular}{@{}c}
\includegraphics[width=0.85\linewidth]{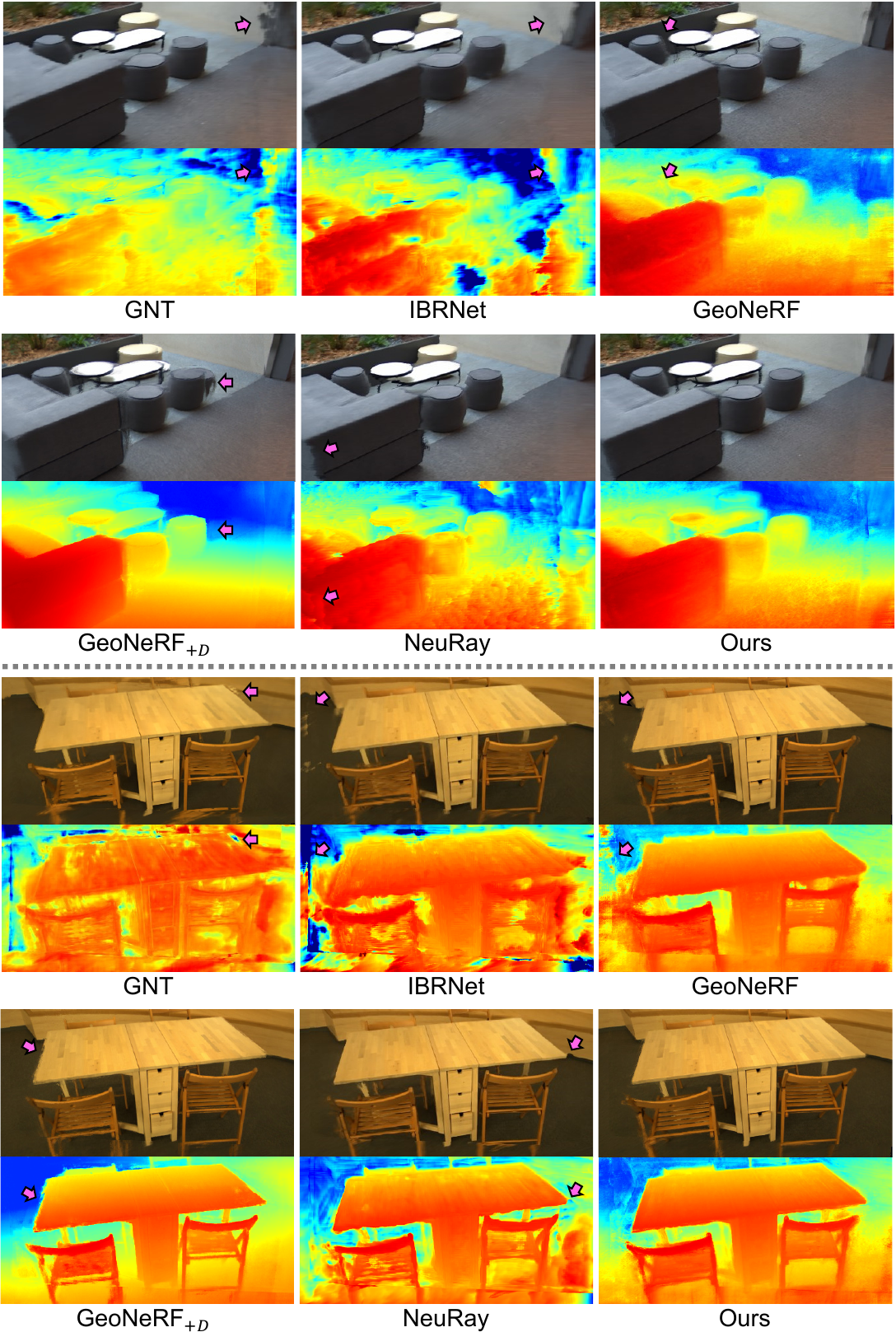} \\
\end{tabular}
\vspace{-3mm}
\caption{
Qualitative comparison of novel-view synthesis on the StereoNVS-Real dataset.
All models are trained using stereo images.
Our method outperforms the baseline methods~\cite{wang2021ibrnet,varma2022attention,johari2022geonerf,liu2022neural} for both image and depth qualities.
}
\label{fig:supple_qual_main}
\vspace{-3mm}
\end{figure*}

\clearpage
\clearpage

{
    \small
    \bibliographystyle{ieeenat_fullname}
    \bibliography{supple}
}
